\documentclass[journal]{IEEEtran}
\usepackage{adjustbox}
\usepackage{array,multirow}
\usepackage[table,xcdraw]{xcolor}
\usepackage{makecell}
\usepackage{float}
\usepackage{comment}
\usepackage{graphicx}
\usepackage{color}
\usepackage{pgfplotstable}
\usepackage{verbatim}
\usepackage{textcomp, gensymb}
\usepackage{amssymb}
\usepackage{rotating}
\usepackage[normalem]{ulem}
\usepackage{gensymb}
\usepackage{pdflscape}
\usepackage{subcaption}   
\usepackage{amsmath}
\usepackage{tabularx}

\usepackage{amssymb}
\usepackage{lineno}
\usepackage{flushend}
\usepackage{adjustbox}
\usepackage{ctable} 
\usepackage[font=small,skip=2pt]{caption}
\usepackage{scalerel}
\usepackage{tikz}
\usepackage{soul}
\usepackage{array} 
\usepackage{amsmath}
\usepackage{algorithm}
\usepackage{algorithm} 
\usepackage{algpseudocode} 

\usepackage{graphicx}
\usepackage{adjustbox}
\usepackage{import}
\usepackage{amssymb}

\usepackage[utf8]{inputenc}
\usepackage{mathtools, nccmath}
\usepackage{amssymb}
\usepackage{pifont}

\usepackage{titlesec}
\titlespacing*{\section}{0pt}{0\baselineskip}{0\baselineskip}
\titlespacing*
    {\subsection}
    {1pt}
    {1ex}
    {1ex}

\makeatletter
\def\@citex[#1]#2{\leavevmode
\let\@citea\@empty
\@cite{\@for\@citeb:=#2\do
{\@citea\def\@citea{,\penalty\@m\ }%
\edef\@citeb{\expandafter\@firstofone\@citeb\@empty}%
\if@filesw\immediate\write\@auxout{\string\citation{\@citeb}}\fi
\@ifundefined{b@\@citeb}{\hbox{\reset@font\bfseries ?}%
\G@refundefinedtrue
\@latex@warning
{Citation `\@citeb' on page \thepage \space undefined}}%
{\@cite@ofmt{\csname b@\@citeb\endcsname}}}}{#1}}
\makeatother
\usetikzlibrary{svg.path}

\definecolor{orcidlogocol}{HTML}{A6CE39}
\tikzset{
  orcidlogo/.pic={
    \fill[orcidlogocol] svg{M256,128c0,70.7-57.3,128-128,128C57.3,256,0,198.7,0,128C0,57.3,57.3,0,128,0C198.7,0,256,57.3,256,128z};
    \fill[white] svg{M86.3,186.2H70.9V79.1h15.4v48.4V186.2z}
                 svg{M108.9,79.1h41.6c39.6,0,57,28.3,57,53.6c0,27.5-21.5,53.6-56.8,53.6h-41.8V79.1z M124.3,172.4h24.5c34.9,0,42.9-26.5,42.9-39.7c0-21.5-13.7-39.7-43.7-39.7h-23.7V172.4z}
                 svg{M88.7,56.8c0,5.5-4.5,10.1-10.1,10.1c-5.6,0-10.1-4.6-10.1-10.1c0-5.6,4.5-10.1,10.1-10.1C84.2,46.7,88.7,51.3,88.7,56.8z};
  }
}

\newcommand\orcidicon[1]{\href{https://orcid.org/#1}{\mbox{\scalerel*{
\begin{tikzpicture}[yscale=-1,transform shape]
\pic{orcidlogo};
\end{tikzpicture}
}{|}}}}
\usepackage{hyperref} 

\ifCLASSINFOpdf

\else
 
\fi

\DeclareUnicodeCharacter{2212}{-}
\begin{document}

\title{ Neuromorphic Vision-based Motion Segmentation with Graph Transformer Neural Network}
\author{

	Yusra~Alkendi$^{1}$ \orcidicon{0000-0001-6618-5317},
	Rana~Azzam$^{2,4}$ \orcidicon{0000-0003-0378-1909},
	Sajid Javed$^{2,5}$ \orcidicon{0000-0002-0036-2875},
    Lakmal~Seneviratne$^{2}$ \orcidicon{0000-0001-6405-8402}, 
	and~Yahya~Zweiri$^{2,3,4}$ \orcidicon{0000-0003-4331-7254}\
 \thanks{$^{1}$Yusra Alkendi is with the Propulsion and Space Research Center (PSRC) at the Technology Innovation Institute (TII), Abu Dhabi, UAE, e-mail: {\small \{Yusra.Alkendi@tii.ae\}}}	
\thanks{$^{2}$Rana~Azzam, Sajid~Javed, and Lakmal~Seneviratne are with the Khalifa University Center for Autonomous Robotic Systems (KUCARS), Khalifa University of Science and Technology, Abu Dhabi, UAE}	
	\thanks{$^{3}$Yahya Zweiri is also with the Advanced Research and Innovation Center (ARIC), Khalifa University of Science and Technology, Abu Dhabi, UAE.}	
	\thanks{$^{4}$Rana Azzam, and Yahya Zweiri are also with the Department of Aerospace Engineering, Khalifa University of Science and Technology, Abu Dhabi, UAE.}

	\thanks{$^{5}$Sajid Javed is also affiliated with the Department of Computer Science, Khalifa University of Science and Technology, Abu Dhabi, UAE.}
  }
\markboth{IEEE TRANSACTIONS ON MULTIMEDIA, 2024}{Alkendi Y. \MakeLowercase{\textit{et al.}}: Bare Demo of IEEEtran.cls for IEEE Journals}
\maketitle

\begin{abstract}
Moving object segmentation is critical to interpret scene dynamics for robotic navigation systems in challenging environments.
Neuromorphic vision sensors are tailored for motion perception due to their asynchronous nature, high temporal resolution, and reduced power consumption. However, their unconventional output requires novel perception paradigms to leverage their spatially sparse  and temporally dense nature. 
In this work, we propose a novel event-based motion segmentation algorithm using a Graph Transformer Neural Network, dubbed GTNN.
Our proposed algorithm processes event streams as 3D graphs by a series of nonlinear transformations to unveil local and global spatiotemporal correlations between events. 
Based on these correlations, events belonging to moving objects are segmented from the background without prior knowledge of the dynamic scene geometry. 
The algorithm is trained on publicly available datasets including MOD, EV-IMO, and \textcolor{black}{EV-IMO2} using the proposed training scheme to facilitate efficient training on extensive datasets. Moreover, we introduce the Dynamic Object Mask-aware Event Labeling (DOMEL) approach for generating approximate ground-truth labels for event-based motion segmentation datasets. 
We use DOMEL to label our own recorded Event dataset for Motion Segmentation (EMS-DOMEL), which we release to the public for further research and benchmarking. Rigorous experiments are conducted on several unseen publicly-available datasets where the results revealed that GTNN outperforms state-of-the-art methods in the presence of dynamic background variations, motion patterns, and multiple dynamic objects with varying sizes and velocities. 
GTNN achieves significant performance gains with an average increase of 9.4\% and 4.5\% in terms of motion segmentation accuracy (\textit{IoU}\%) and detection rate (\textit{DR}\%), respectively.


\end{abstract}

\begin{IEEEkeywords}
Neuromorphic Vision, Dynamic Vision Sensor, Event Camera, Motion Segmentation, Graph Transformer Neural Networks.
\end{IEEEkeywords}

\IEEEpeerreviewmaketitle
 
\section{Introduction}{
Scene understanding constitutes a cornerstone to a plethora of robotic applications where automation and behavioral intelligence are essential, such as navigation \cite{desouza2002vision,alkendi2021state}, exploration \cite{exploration}, and simultaneous localization and mapping \cite{liang2018salientdso}. 
It incorporates perceiving sensory information to infer geometric and semantic properties of objects present in the context of a working environment.
The robustness of such perception modules heavily depends on their ability to cope with various inherent environmental challenges, such as scene dynamics, varying lighting conditions, and the absence of prior knowledge related to the number of objects. 
According to \cite{rateke2022passive,beghdadi2022comprehensive,bi2022survey}, approaches that can tackle such challenges are still under investigation.
\textcolor{black}{Scene Segmentation can be broadly divided into three main categories: Semantic \cite{add4_2023}, Instance \cite{ADD2_2019, ADD3_2020}, and Motion Segmentation \cite{add5_2020}. A recent progress in this area can be contributed towards advanced deep learning approaches exhaustively reviewed in \cite{ADD2_2022}. Depending on specific application requirements, appropriate segmentation techniques can be selectively employed for accurate scene analysis. }
In a navigation scenario, where a robotic vehicle moves in its task environment and is anticipated to encounter one or more dynamic objects, motion segmentation is pivotal to ensure safety of the vehicle and its surrounding environment, and hence ascertain the continuity of the robotic task. 
Motion segmentation is defined as the segregation of \textcolor{black}{some dynamic object}, moving independently, from the background motion, based on observations acquired from passive \cite{mattheus2020review} or active \cite{activeMS} sensors. 

\begin{figure}[]{
\centering
\setlength{\fboxrule}{1.pt}%
\centering
\begin{adjustbox}{width=0.46\textwidth}
  \centering
  {\Large
\begin{tabular}{ccc}

 \fbox{\includegraphics[width=\columnwidth,height=7.1cm]{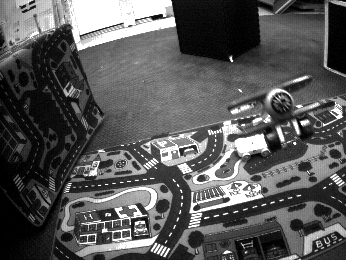}}\centering
&\fbox{\includegraphics[width=\columnwidth,trim={3.8cm 2.4cm 2.8cm 1.5cm},clip]{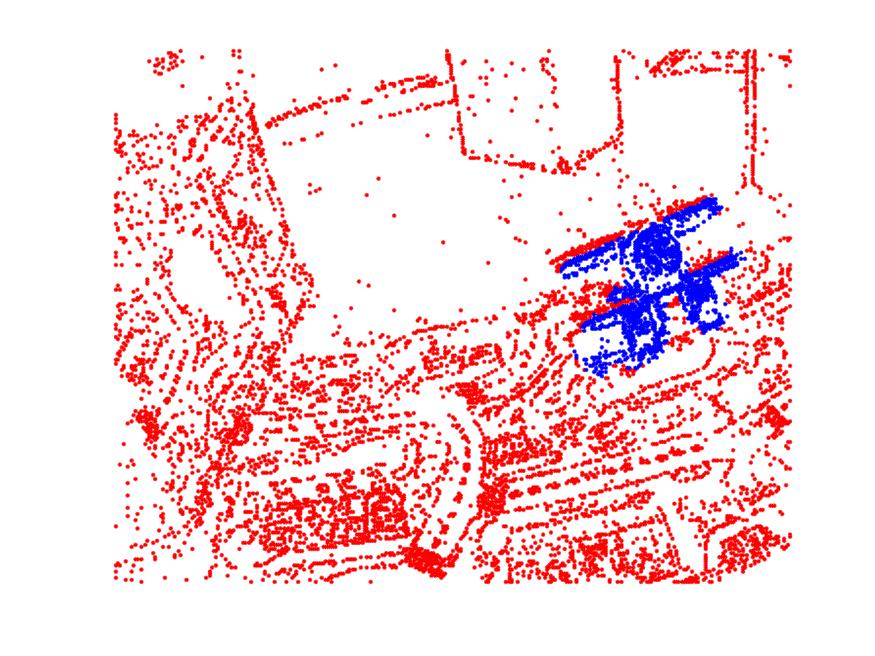}}\centering
&\fbox{\includegraphics[width=\columnwidth,trim={3.8cm 2.4cm 2.8cm 1.5cm},clip]{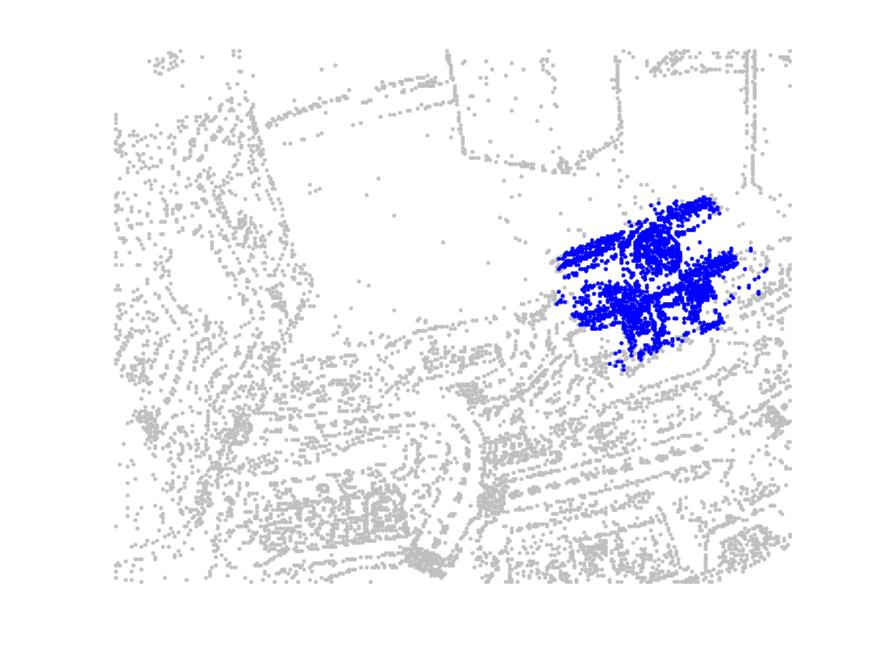}}
\\
\Huge{\begin{tabular}{c} (A)\end{tabular}} &\Huge{(B)}& \Huge{(C)} \\
\end{tabular}}
\end{adjustbox}
 \setlength{\belowcaptionskip}{-8pt}   
\caption{Motion segmentation results of the proposed learning-based algorithm (GTNN) using EV-IMO (Floor sequence) publicly available dataset \cite{m6_evimo}. (A) APS image for visualization only. (B) Approximate Ground truth events: \textcolor{black}{black} represents foreground events and \textcolor{black}{red} represents background events. (C) Segmented events using the proposed GTNN algorithm: \textcolor{black}{black} indicates predicted foreground events and \textcolor{black}{gray} indicates predicted background events. 
Our GTNN performs a binary classification to differentiate between foreground events due to moving objects or background events due to camera motion.}\label{into_example}}\end{figure}

Vision-based motion segmentation is an active research area and several approaches have been proposed in the literature \cite{Event_based_Vision_A_Survey}. 
Neuromorphic vision is an emerging research area and technology that mimics the working principle of the human retina, by timely and asynchronously capturing the polarity of the log intensity variations in the observed scene at the pixel level. 
Neuromorphic vision sensors are also referred to as event-based cameras\footnote{These two terms (Neuromorphic vision sensors and event-based cameras) will be used interchangeably throughout the paper.} acquire information in the form of continuous streams at low latency ($>$$20\mu$$s$), high temporal resolution ($>$800kHz), high dynamic range ($>$120 dB), and no motion blur \cite{Event_based_Vision_A_Survey}.
Event-based cameras outperform the capabilities of conventional cameras and therefore make them an ideal alternative for conventional vision sensors, particularly in applications that require robust perception, such as motion estimation of vehicles under varying illumination conditions in the presence of dynamic scenes. 
A wealth of event-based computer vision literature has unlocked new opportunities and initiated research directions for more robust visual perception. 
Such capabilities have been demonstrated in various applications such as visual localization \cite{s27_UltimateSlam,salah2022neuromorphic}, depth estimation \cite{furmonas2022analytical}, object tracking \cite{jiang2020object}, precision manufacturing \cite{ayyad2023neuromorphic}, and tactile sensing \cite{macdonald2022neuromorphic}, to name a few. 
Though event cameras have demonstrated significant advantages over their conventional counterparts, respective computer vision paradigms must handle the data sparsity and low spatial resolution typical of the acquired event streams.

Recently, few research work are conducted to develop motion segmentation paradigms based on data acquired from event cameras.
These paradigms can be classified into: (1) classical approaches such as \cite{stoffregen2019eventbased,M10_Zhou21tnnls,M8}, and (2) learning-based approaches such as \cite{SpikeMS,mitrokhin2020learning}. 
In the former, motion-model fitting and clustering methods are used to differentiate the events that belong to a moving object in the scene from those belonging to the background. 
In the latter, classification or regression using neural networks, such as spike and graph convolutional, are employed to identify events belonging to a moving object. 
These approaches either require prior knowledge of the scene (number of objects) and camera motion, camera parameter tuning, sensor data pre-processing (filtering, offline parameter computation, etc.), or initialization of the optimization problem. 
These limitations hinder the generality and applicability of the state-of-the-art (SOTA) event-based segmentation approaches and hence, this problem remains unsolved. 

In the current work, we propose a learning-based motion segmentation algorithm\footnote{A supplementary video is available at: $<$\url{https://youtu.be/rVvbKdXh6oE}$>$} to aid robotic navigation in unknown dynamic environments. 
Particularly, we employ a Graph Transformer Neural Network (GTNN), based on the point transformer layer \cite{Point_transformer}, to classify events triggered in a dynamic scene into moving-object events or background events. Our GTNN approach is based on processing raw event streams, acquired by an event camera, without requiring any prior knowledge about the topology or the dynamics in the scene, prior knowledge of the camera motion, camera parameter tuning, event pre-processing, or offline initialization.
Eliminating such requirements is essential to achieve better generalizability across various scenes, motion scenarios, and sensors.

Transformers have been recently used for event-based applications and have demonstrated outstanding performance in processing event streams for event-denoising \cite{AlkendiY} object detection \cite{TransformersDavid}, and action and gesture recognition \cite{EventTransformer}.
However, it is worth noting that the transformer operates on 2D frames in \cite{EventTransformer,TransformersDavid}. 
These frames were reconstructed from events, as opposed to the work in \cite{AlkendiY} where a graph-driven transformer operates on raw event data. 
\textcolor{black}{A point transformer has recently been proposed to handle dense representations of point clouds for object classification and segmentation using 3D data \cite{Point_transformer}. 
In the current work, we demonstrate the scalability of GTNN using the point transformer layer and address the challenges posed by unconventional and noisy sensor output, as well as the low-spatial resolution and sparsity of neuromorphic vision sensors. More specifically, working with event streams is deemed challenging due to the limited features encoded in raw event data, such as the spatial coordinate of the pixel where the intensity change occurred in the scene, the time of the event, and the polarity; +1 or -1 indicating an increase or decrease in intensity, respectively.} \textcolor{black}{To handle the differences in input data between the sparse representation of the event stream and the dense representation of a point cloud, a new architecture for the GTNN algorithm has been developed. This architecture includes a point transformer layer, as well as some additional changes in the number of encoder and decoder units and the addition of processing layers such as global aggregation. While the direct implementation of the point transformer structure developed by \cite{Point_transformer} was initially considered, evaluations have shown that it was too complex and less efficient for the event-based motion segmentation task at hand. } 

The proposed GTNN algorithm is trained in a supervised manner on publicly available datasets \cite{m6_evimo,M8,EVIMO2} to perform end-to-end motion segmentation, i.e GTNN takes as input unprocessed event streams and outputs the segmentation results. 
Evaluation sequences from the publicly available datasets are used to validate the proposed algorithm and to quantitatively and qualitatively compare it against the SOTA learning-based motion segmentation approach \cite{SpikeMS} and the offline classical approaches \cite{M10_Zhou21tnnls,M8,M9_s128_Mitrokhin_2018}.
The segmentation accuracy achieved by GTNN outperforms SOTA motion segmentation approaches \cite{SpikeMS,M10_Zhou21tnnls,M8,M9_s128_Mitrokhin_2018}.
Further to that, the GTNN model is more efficient than \cite{SpikeMS} in terms of computational requirements and hence performs faster prediction.
The testing scenarios exhibited various scene dynamics, changing illumination conditions, and different camera motion dynamics. 
The proposed algorithm is robust against all of the aforementioned challenges and requires no fine-tuning upon testing in unknown environments using unseen sequences.
Fig. \ref{into_example} shows sample segmentation results obtained when testing our proposed algorithm on a publicly available dataset of EV-IMO (Floor) \cite{m6_evimo}.

Additional testing sequences are recorded in our lab facilities to further analyze the performance of the proposed algorithm in scenarios recorded in a different domain than that of the publicly available training and testing datasets. 
To that end, we propose the \textcolor{black}{Dynamic Object} Mask-aware Event Labeling (DOMEL) approach to generate approximate ground-truth event labels for the recorded sequences. 
The performance of GTNN on these recorded sequences has verified its generalization across new domains. 

The overall framework proposed in this paper is depicted in Fig. \ref{fig:dvs_proposedApproach1} highlighting the detailed design of (1) the graph transformer neural network (GTNN) for motion segmentation and (2) the dynamic object mask-aware event labeling (DOMEL).

To summarize, the contributions of this work are:

\begin{enumerate}
\item The design and development of an event-based motion segmentation algorithm, based on GTNN.
The proposed GTNN (1) preserves the asynchronous nature of event streams and exploits spatiotemporal correlations to infer the scene and camera motion dynamics, (2) does not require any prior knowledge about the scene geometry and/or dynamics, and (3) does not need initialization or event pre-processing to perform motion segmentation. 

\item The design of an effective training scheme that facilitates training on an extensive dataset while reducing computational requirements. 
This scheme enables faster convergence and hence higher performance than conventional training by exposing portions of the training data to the network at every iteration.

\item Extensive evaluation of the proposed GTNN algorithm on publicly available event datasets and on experimental sequences recorded locally in our lab facilities. 
Comparisons to the SOTA learning-based and classical motion segmentation approaches have also shown the superiority of the proposed algorithm in terms of segmentation accuracy (\textit{IoU}\%) and detection rate (\textit{DR}\%). 

 \item The release of a new event dataset (EMS-DOMEL) with the corresponding motion segmentation ground truth labels obtained using the proposed DOMEL approach. 
 Data is recorded using two event cameras with different resolutions, in scenes involving multiple dynamic objects of various sizes/types in a variety of challenging environmental scenarios. 

 The dataset can be accessed through the following link: $<$\url{https://github.com/Yusra-alkendi/EMS-GTNN}$>$ for further research and benchmarking.

\end{enumerate}

\noindent The remainder of this paper is organized as follows.
Section \ref{sec:related-work} provides a review of the related literature. 
Section \ref{sec:proposedframework} describes in detail the proposed GTNN algorithm and the EMS-DOMEL dataset. 
In Section \ref{sec:results}, the experimental results are presented, analyzed, and compared to SOTA motion segmentation approaches. The conclusions of the current work are drawn in Section \ref{sec:conclusion}.

\begin{figure*}[]{
\centering
 \includegraphics[width=0.88\textwidth]{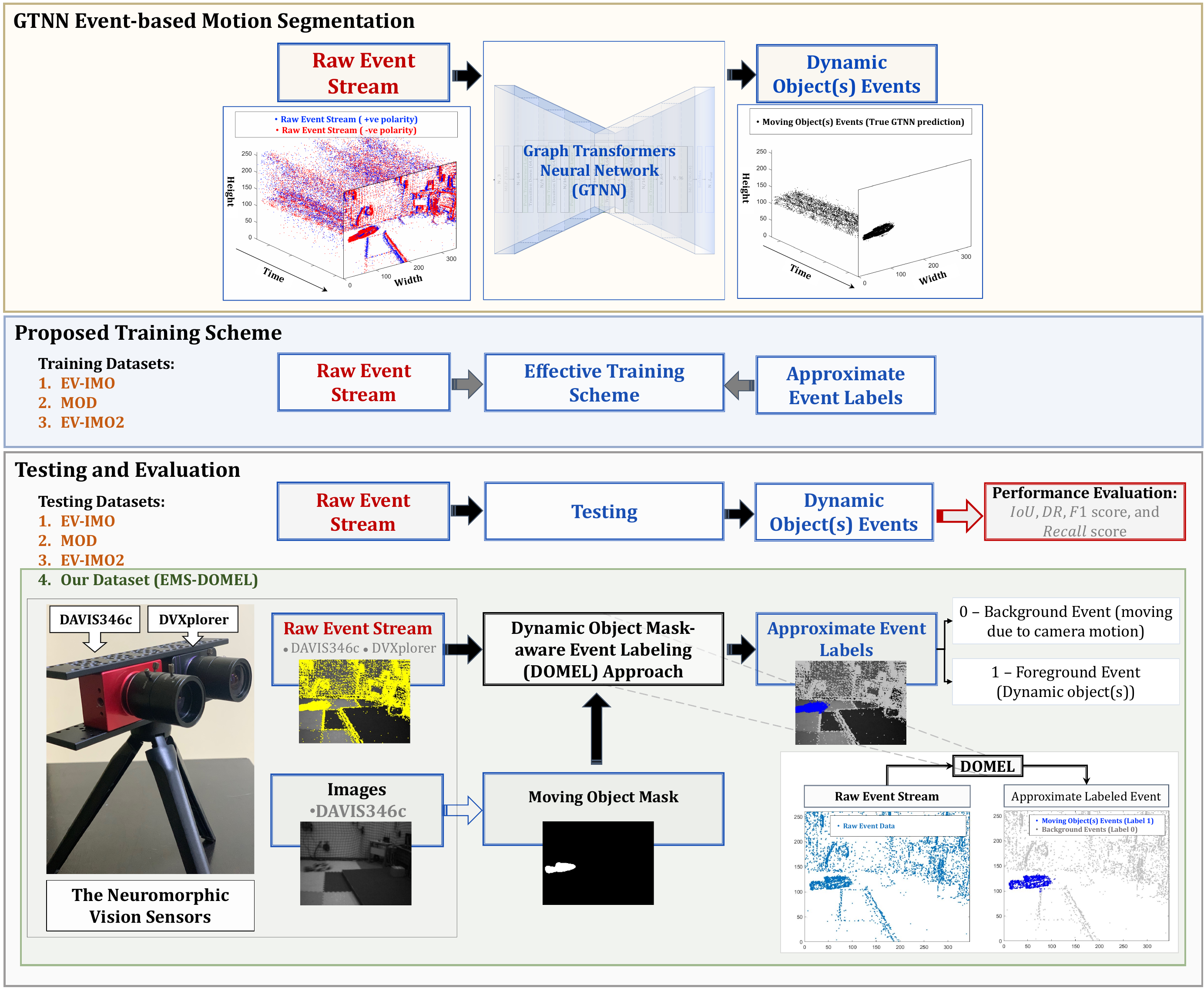}
 \setlength{\belowcaptionskip}{-8pt}   

  \caption{Proposed event-based motion segmentation framework based on graph transformer neural network (GTNN). GTNN is developed and trained using the proposed training scheme on publicly available event datasets (EV-IMO, MOD, EV-IMO2) and tested on the corresponding evaluation sequences along with our recorded experiments (EMS-DOMEL). The approximate ground truth event labels of our recorded experiments are generated using the proposed DOMEL approach. The proposed algorithm classifies incoming event streams into foreground events related to moving object(s) or background events.
  }
  \label{fig:dvs_proposedApproach1}}
\end{figure*}}

\section{Background and Related Works} \label{sec:related-work}
\subsection{Event-based Motion Segmentation}
Solutions for event-based motion segmentation have been investigated throughout the past years for variable environmental conditions at different complexity levels. Owing to the camera's working principle, events are only generated if changes occur, either due to camera motion or dynamics in the scene. For the simplest scenario where the event camera is static, motion segmentation has been tackled using clustering approaches as in \cite{4_barranco2018realtime,20_etracking,23_Saccade}.

In case the camera capturing the scene is moving, events are generated due to two main reasons: (1) camera ego-motion, and (2) moving objects. To differentiate between these two categories, camera motion should be estimated and/or additional information about the environment is required \cite{29_occlusions}.
A recent motion segmentation framework based on motion compensation has been proposed in \cite{stoffregen2019eventbased}, where raw events are accumulated into 2D frames. The frames are then aligned with high contrast contours, and evaluated using dispersion \cite{M9_s128_Mitrokhin_2018} or sharpness \cite{s124_gallego2019focus,26_sharpness} measures.
For high dynamic environments, Mitrokhin et al. \cite{M9_s128_Mitrokhin_2018} have proposed an approach to detect moving objects by fitting a motion model using multi-key parameters into the background, and then considering misaligned events to belong to different segments/classes. The clustered objects are then tracked by a Kalman filter to handle occlusions and scene uncertainties. This approach might not work if multiple moving objects appear in one scene.
This work was later extended in \cite{stoffregen2019eventbased} by integrating the Expectation-Maximization (EM) approach in the segmentation method. Their results showed excellent segmentation and optical flow estimation. However, the EM-based segmentation algorithm requires prior knowledge about the scene, particularly the number of moving objects to start initialization. 

Recently, Parameshwara et al. \cite{M8}, have proposed an approach using a combined nonlinear optimization method to segment multiple objects moving independently, without knowledge of the number of moving objects in the scene. Furthermore, Mitrokhin et al. \cite{m6_evimo} proposed a learning-based framework for motion segmentation using motion compensation, where depth, ego-motion, and clustering of independent moving objects (IMOs) and their 3D velocity were estimated. 

A recent study by Zhou et al. \cite{M10_Zhou21tnnls} developed an offline optimization approach based on energy minimization where identification of IMOs acquired with an event-based camera is performed based on motion fitting and clustering methods. The iterative scheme allows for exploiting spatiotemporal correlations between event streams for event identification. The proposed algorithm requires no prior information about the scene, dynamic objects within the scene, or camera motion. It, however, requires an initialization stage to start the optimization. 

Lastly, Parameshwara et al. \cite{SpikeMS} developed a learning-based motion segmentation approach using an encoder-decoder spike neural network architecture, called a SpikeMS model. SpikeMS is a binary classifier based on a novel spatiotemporal loss function. SpikeMS is capable of performing incremental predictions for event streams where the segmentation accuracy is comparable to offline classical-based approaches \cite{M9_s128_Mitrokhin_2018,stoffregen2019eventbased,M8}.  
Spatiotemporal correlation methods and deep learning approaches have shown potential in the reviewed event-based motion segmentation approaches, however, various aspects of this field remain largely unexplored.

\subsection{Graph Transformers Neural Networks}

Deep learning models which operate on non-Euclidean graph-structured data are known as graph neural networks (GNNs). GNNs have been successfully employed in a multitude of applications \cite{gnn_review} such as object segmentation \cite{giraldo2020graph} due to their expressive power and modeling flexibility. GNNs perform mapping of input graphs based on their node features and connectivity within a neighborhood (edges), despite the order in which they are fed to the neural network. It is also worth noting a single GNN architecture may accept input graphs of variable sizes. This important property of GNNs highly suits the asynchronous nature and variable input rate of event streams, which are a function of the scene and camera motion dynamics. 

Transformers, on the other hand, have recently proven cutting-edge performance in a variety of applications, including natural language processing \cite{vaswani2017attention} and computer vision tasks \cite{han2020survey, TMM_paper1, TMM_paper2, TMM_paper3, TMM_paper4}. 
The self-attention head used in transformers is responsible for capturing the relationships between inputs and outputs and allowing simultaneous processing of sequential recurrent networks. Graph-based transformers with their self-attention operation have proven an outstanding capability for processing 3D data such as 3D point cloud \cite{Pointcloudtransformer,Point_transformer,pointyu2022point,pointhan2022dual} for vision tasks like segmentation and classifications. Inspired by this success in 3D data processing and our previous work on GNN-transformer for events denoising \cite{AlkendiY}, we design a learning-based motion segmentation network based on a graph transformer neural network. The proposed algorithm allows for handling the asynchronous nature of events for revealing their spatiotemporal correlations and processing the scene dynamics accordingly. An event stream is fed to the algorithm, which extracts the local features of every event and integrates them with a global feature representing the whole stream, prior to processing them using nonlinear operations. The event stream is then segmented into events that belong to the background and others that belong to moving objects in the scene.

\section{Proposed Framework}\label{sec:proposedframework}{

In this section, the proposed design of (1) the graph transformer neural network (GTNN) for motion segmentation and (2) the dynamic object mask-aware event labeling \textcolor{black}{(DOMEL)} are presented in detail.

\subsection{Graph Transformer Neural Network (GTNN) Algorithm for Event-based Motion Segmentation}\label{sec:proposedGTNN}

The GTNN motion segmentation algorithm operates on incoming event streams, acquired through a moving event camera in a dynamic scene to carry out classification of events into (1) foreground events: events that belong to dynamic objects in the scene, and (2) background events: events that belong to the static background and were generated due to camera ego-motion. GTNN processes events in their raw \textcolor{black}{formats}, i.e. does not perform any event pre-processing such as accumulation into 2D frames, and hence preserves their asynchronous nature. Particularly, events are structured as 3D graphs encapsulating their spatiotemporal properties. Event graphs are then passed through a set of convolution and deconvolution operations as per the architecture visualized in Fig. \ref{fig:dvs_network}. 

The remainder of this section presents in great detail the implementation of the proposed GTNN algorithm. Section \ref{sec:graph_construction} provides details on \textcolor{black}{events-3D} graph construction based on k-nearest neighborhood ($k$NN) strategy, Section \ref {sec:event_graph_transformations} explains the \textcolor{black}{events-3D} graph transformation process where the graph nodes and their features are nonlinearly mapped by the encoding-decoding GTNN model, and Section \ref{sec:gtnn_architecture} delineates the architecture of the proposed GTNN. 

\begin{figure*}[!bt]
\centering
 \includegraphics[width=0.87\textwidth]{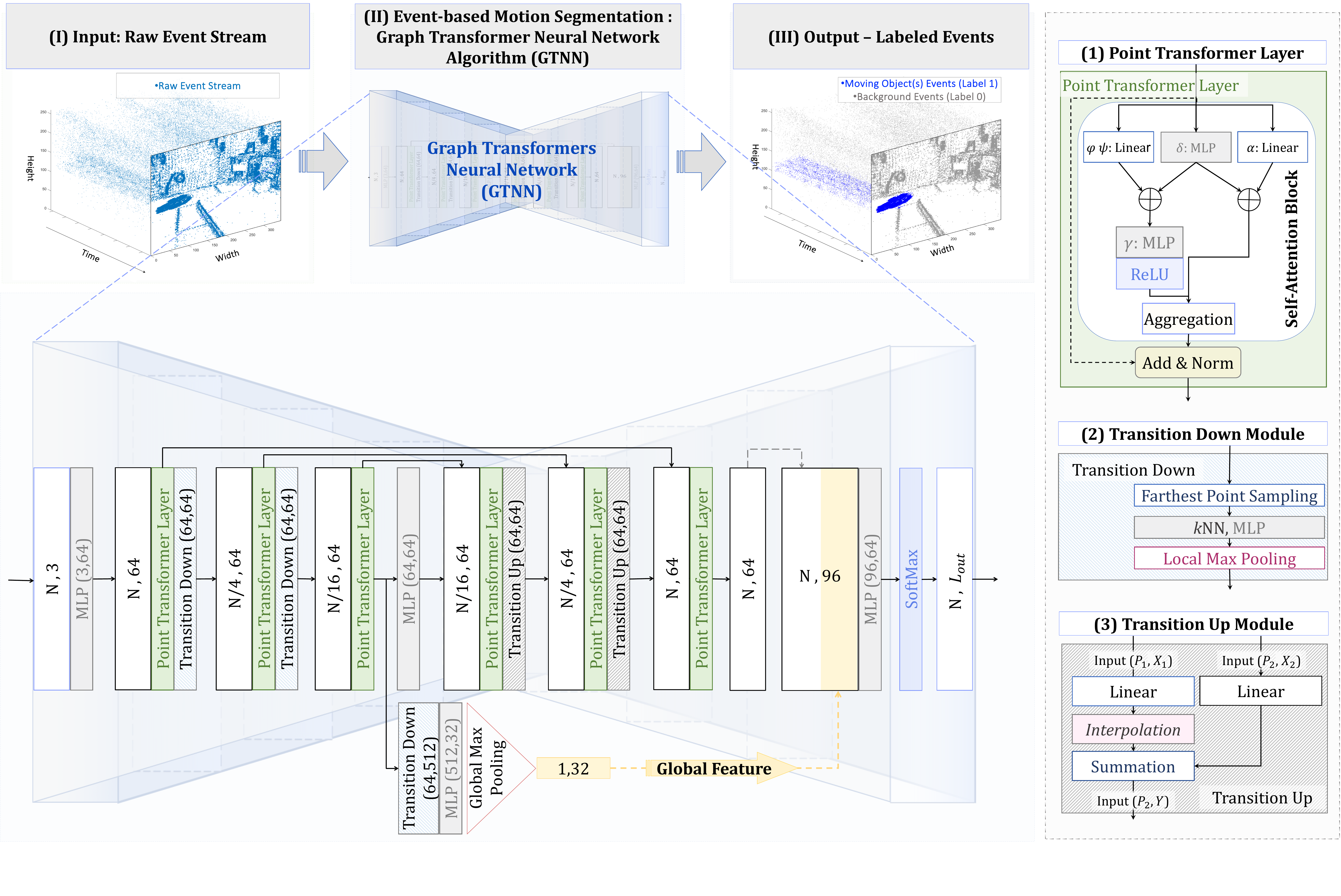}
 \setlength{\belowcaptionskip}{-12pt}   

  \caption{Proposed Graph Transformer Neural Network (GTNN) Architecture for Event-based Motion Segmentation.
  The classification network takes $N$ number of events as input stream, applies input and feature transformations and mapping, and then aggregates global features by global max pooling. The output is a \textcolor{black}{binary classification assigned to each event in the stream, indicating whether it belongs to class 0 (a moving background due to camera motion) or class 1 (dynamic object(s) within the scene).}}
  
  \label{fig:dvs_network}
\end{figure*}

\subsubsection{Events-3D graph construction} \label{sec:graph_construction}

Data perceived by an event camera report changes in log intensities of the observed scene in the form of asynchronous events. Events span a spatial resolution of $H$$\times$$W$ pixels, where $H$ and $W$ are the vertical and horizontal dimensions of the camera's frame. A stream of $N$ events, $\{e_i\}_N$, can be represented using a sequence of 4-tuples as indicated below: 
\begin{equation}
{\{e_i\}_N =\{x_i, y_i, t_i, p_i\}_N\textcolor{black}{,}}
\end{equation}

\noindent where ($x_i$, $y_i$) 
\textcolor{black}{is the event’s pixel coordinate} at which event $i$ has occurred within the frame, $t_i$ refers to the time at which event $i$ is triggered, and $p_i$ is the event's polarity. Event polarity may take one of two values\textcolor{black}{:} $+1$ if the brightness of the pixel has increased and $-1$ otherwise.

A stream of events triggered within a pre-defined temporal window is structured as a 3D graph $G$. In case of aggressive vehicle maneuvers, and due to the high temporal resolution of the event camera, a huge amount of event pixels will be active simultaneously. Consequently, inferring spatiotemporal event relationships is deemed challenging due to data redundancy and high computational requirements. To resolve this issue and to minimize memory usage, the size of the graph constructed from the temporal window is limited to a fixed maximum ($N_{max}$=5000), which was decided based on visual assessments of the events' projection on a 2D frame. 
This means that the temporal window could be shorter if too many events were received as a result of high scene dynamics, in which case the most recent $N_{max}$ events will be used to construct the graph.
As for memory requirements, classification of an event sample requires storing up to $N_{max}$ events. It is worth noting that graphs constructed from a single event sequence may exhibit variable sizes at different times, depending on the scene and camera dynamics. Operation on variable-sized graphs is a property of graph-based neural networks. By virtue of this, our proposed approach may generalize across various domains, regardless of the number of triggered events in the temporal window.

Each node in the graph $G$ represents an event in the stream. Consequently, node features are the properties of the triggered events, namely $<$$(x_j), (y_j), (t_j)$$>$, where $j$ represents the node index in the graph, ($x_j$, $y_j$) are the event's pixel coordinates, and $t_j$ is the event's timestamp. Although event polarities may provide insights about the motion direction through analysis of brightness changes, this property is sensitive to the camera's parameter settings. To that end, performing motion segmentation based on this event property may limit the generality of the proposed algorithm, hence why we omitted it from the graph node features. A similar consideration was adopted in our previous work \cite{AlkendiY}. 

Next, the $k$-Nearest Neighbor ($k$NN) search is performed to connect every node ($e_i$) to k-nearest neighboring nodes ($e_j$) based on their 3D spatiotemporal distances. Resulting spatiotemporal neighborhoods are referred to as sub-graphs, each containing $k+1$ nodes. These sub-graphs will be further processed by the encoding-decoding operations as will be discussed in Section \ref{sec:event_graph_transformations}.

\subsubsection{\textcolor{black}{Events-3D} graph transformations}\label{sec:event_graph_transformations}

The constructed 3D graph, including all the sub-graphs, will be passed through a set of encoding and decoding layers in GTNN. In this section, the details of the core layers will be explained; namely a point transformer layer, transition down module, and transition up module. 

\textbf{Point Transformer Layer:} 
The point transformer layer operates on the sub-graphs generated in the previous step. It is worth mentioning that several implementations of transformer layers for classification and segmentation of 3D data were found in the literature, such as \cite{Point_transformer,Pointcloudtransformer}. The implementation adopted in our proposed approach was inspired by \cite{Point_transformer}, since the constructed 3D event graphs exhibit the same structure as 3D point clouds, although the data inference is different.


The point transformer layer implements a self-attention mechanism that captures the sequence relationship between the inputs and outputs for structured prediction tasks. The attention mechanism based on the query–key–value (QKV) model enables functions with high long-term memory \cite{vaswani2017attention} and executes parallel processing which can reveal jointly complex relationships between inputs and outputs. The most popular self-attention operators are scalar attention \cite{vaswani2017attention} and vector attention \cite{vector_attention}. 
The vector or dot-product attention operator (i.e. a simple matrix multiplication) is selected in our algorithm to achieve faster state update and better space efficiency. 
The point transformer layer is composed of a residual block to perform feature aggregation and feature transformation as illustrated in Fig. \ref{fig:dvs_network}-(1).

Let $X(i)$ be a set of input feature vectors. $X(i)$ will be processed by three connected operation streams. In the first stream operation, a subtraction relation is used between the input features after being transformed in a point-wise manner by $\varphi$ and $\psi$. Those will be added with a position encoding $\delta$ (from the second operation) and consequently forming the attention vector which is nonlinearly mapped by $\gamma$ through MLP. 
Concurrently, in the third operation stream, input features are transformed by $\alpha$ and added to a position encoding $\delta$. Then, the outputs of all stream operation lines are aggregated using the Hadamard product. $\varphi$, $\psi$, and $\alpha$ are point-wise feature transformations, such as linear projections or nonlinear MLPs. $\delta$ is a position encoding function. $\gamma$ is a mapping function (in our case, two layers of MLP, followed by ReLU). The output of the point transformer layer is represented by the following formula: 
\begin{equation}
    {y_i = \sum_{x_j \subseteq X(i)}{\rho(\gamma(\varphi(x_i)-\psi(x_j)+\delta)) \odot
 (\alpha(x_j)+\delta)}\textcolor{black}{,}}
\end{equation}

\noindent where $y_i$ is the output feature. $\rho$ is a normalization function. 
Spatiotemporal patterns between a set of events are essential to carry out event-based motion segmentation. Such patterns could be obtained based on extracted local and global feature correlations between events in a graph. By operating on sub-graphs, the point transformer layer will extract information on the local coherence between the events. This local graph operated-attention transformer has been adopted in literature for image analysis \cite{vector_attention} and 3D point cloud processing \cite{Point_transformer}. 

The 3D event graph $G$ is passed to the point transformer layer, where every sub-graph is processed to nonlinearly map the node features. Sub-graphs are referred to as $X(i)$, where $X(i) \subseteq X$, and $X$ is the 3D graph $G$ in our implementation.
The output of the layer is of the same structure as the input graph $G$. Details on the point transformer layer are depicted in Fig. \ref{fig:dvs_network}-(1).

\textbf{Transition Down Module:} In this module, the cardinality of the 3D graph is reduced by a certain factor, to convolve the graph nodes. For instance, for a graph $G$ with $N$ nodes and a reduction factor 4 is requested, the transition down module will output a new graph containing $N/4$ nodes. 
As schematically illustrated in Fig. \ref{fig:dvs_network}-(2), $G(p1)$ refers to the input graph with $p1$ nodes that enters the transition down module where $G(p2)$ is the output graph (reduced graph size) with $p2$ nodes. The farthest point sampling (FPS) algorithm is adopted and performed in $G(p1)$ to identify a well-spread downsample subset $G(p2)$ (where $G(p2)$$\subset$$G(p1)$) with the requested cardinality. 
It is worth noting that the pooling is performed on $G(p1)$ using the same $k$NN graph strategy to obtain $G(p2)$, which is the same neighboring set of data previously identified in the point transformer layer. 
In this work, $k$ is selected to be 16 based on the ablation study which is performed on this hyperparameter and its variants. In this module, each input feature is passed sequentially through a linear transformation, batch normalization, and ReLU operations. Then a max pooling is operated onto each node in $G(p2)$ based on $k$ neighborhood nodes in $G(p1)$.

\textbf{Transition Up Module:} 
U-net architecture is adopted  \cite{Point_transformer} where the encoder layers, in our case point transformer layer, are coupled with the corresponding decoder layers. This coupling was adopted since motion segmentation is similar to semantic segmentation which requires a dense prediction-masked output.
The main function of the transition up module is to map features from the reduced graph data set, $G'(p2)$, onto its superset graph data set, $G'(p1)$ (where $G'(p1)$$\supset$$G'(p2)$). Similar to transition down operation, each input point feature is sequentially processed by a linear layer, batch normalization, and ReLU operations. The node features of $G'(p2)$ are mapped to a high-dimension graph size, $G'(p1)$, using trilinear interpolation. To that end, the interpolated features of $G'(p1)$ and the features of corresponding encoder stage $G'(p1)$ are concatenated via the skip connection. In other words, and as mentioned earlier, the point transformer layers operate as the network encoder layers where their output graph is connected to the output graph from transition up modules (decoders) via skip connection. The structure of the transition up module is depicted in Fig. \ref{fig:dvs_network}-(3). 


\subsubsection{The GTNN architecture}
\label{sec:gtnn_architecture}

Fig. \ref{fig:dvs_network} shows the overall architecture of GTNN, where given an input 3D graph, node features will be processed by means of various nonlinear operations to perform motion segmentation. In other words, every event (i.e. node) in the graph will be classified as a foreground event or a background event. 


The event graph is encoded in two stages, as depicted in Fig. \ref{fig:encoder_operation}; (1) node features are encoded using the point transformer layers and (2) graph nodes are encoded using the transition down module. A point transformer layer coupled with a transition down module is referred to as an encoder unit. Similarly, a point transformer layer coupled with a transition up module is referred to as a decoder unit.
\begin{figure}[!t]
\centering
 \includegraphics[width=0.4700\textwidth]{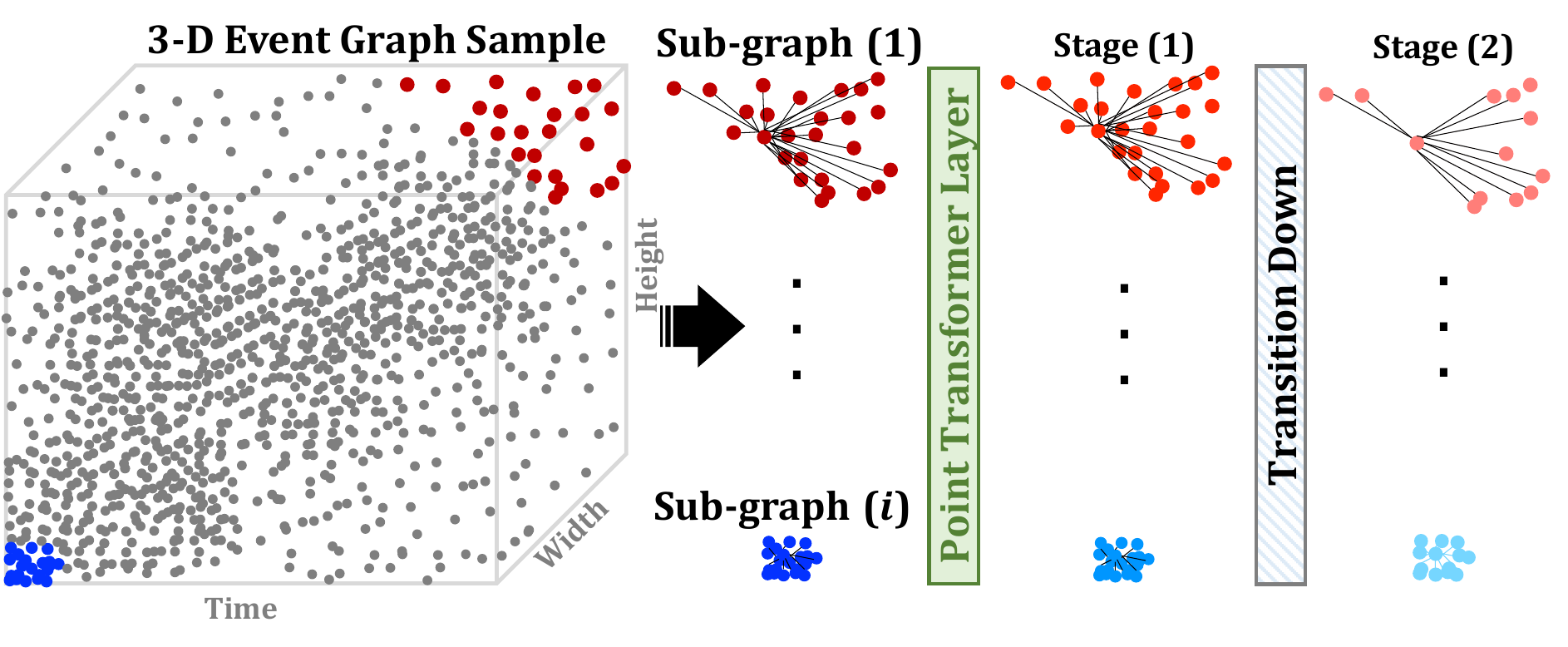}
  \setlength{\belowcaptionskip}{-010pt}   

 \caption{Encoding operation of a sample 3-D event graph by GTNN encoder unit, a coupled point transformer layer (stage 1) and a transition down module (stage 2).}
  \label{fig:encoder_operation}
\end{figure}
\raggedbottom

Based on the application and dynamics in the data, the number of encoder units is varied.
In our case and according to the ablation study, the best-performing network architecture has three encoder-decoder units. The selected downsampling rates for the transition down modules are [1, 4, 4] which correspond to graph sizes of [N, N/4, N/16], where $N$ is the number of events within the graph $G$. Consequently, the upsampling rates for the corresponding transition up modules will be [4, 4, 1]. Hence, the output graph will be of the same size as the input graph. 

Obtaining a global feature vector that correlates the events in the graph is necessary for motion segmentation. This is achieved by transitioning down the graph nodes, obtained after the three encoder layers, into a single node. The features of this node are passed through an MLP and transformed by a graph average pooling operation forming a global feature vector. Inspired by the other segmentation models \cite{Pointcloudtransformer}, the global feature vector is appended to every node feature in the graph after the last decoder layer in the network. 
The output is then passed to an MLP 
followed by a \textcolor{black}{softmax (Eq. (\ref{eq:softmax}))} layer that generates a $N$ × 2 tensor. The entries in each row are the predicted probabilities of an event being a foreground or a background event.

\begin{equation}\label{eq:softmax}{\textbf{Softmax}(x_i)=  \frac{e^{x_i}} {\sum_{j=1}^2 e^{x_j}}} 
\end{equation}
\useshortskip 


A detailed ablation study is presented in the supplementary material where the number of encoder-decoder units was varied, and the performance of the network with and without aggregation of the global feature vector was analyzed. 



\subsection{Proposed Effective Training Scheme}
In our work, a neural classifier is trained on set of 3D graphs, representative of event streams to perform motion segmentation in a supervised manner. The network will carry out node classification, where it will predict a label/class for every node in the graph. Given an input 3D graph and the true node labels, the weights of the classification neural network are optimized to minimize the difference between the predicted output and the ground truth label. The sparsity and the limited information obtained from event data, i.e. $<x_i, y_i, t_i, p_i>$, make scene interpretation and hence, end-to-end vision tasks such as motion segmentation challenging. On the plus side, the high temporal resolution of event data makes it possible to obtain large amounts of data in a short period of time.

The accuracy of deep learning models heavily depends on the quality and quantity of the datasets used for training and evaluation. Exposing neural networks to large amounts of diverse training data is essential for (1) effectively tuning the network's parameters to generate accurate predictions, and (2) enhancing the network's ability to generalize well to unseen samples. Nevertheless, extensive efforts are required to obtain, for some applications, and correctly annotate large amounts of training data. Moreover, operating on large training datasets is computationally expensive and models may not seamlessly converge to the global optimum solution. To circumvent these challenges, researchers have investigated various techniques \cite{optimizie_training1,optimizie_training2} to optimize the selection of training data to improve training efficiency.

In this work, we propose an effective training scheme, to exploit the availability of huge amounts of training data to facilitate global convergence, while reducing the training computational requirements, as shown in Fig. \ref{fig:effective_regular_training_strategy} \textcolor{black}{and Algorithms \ref{alogorithm1}.} 
Instead of using the full training dataset every epoch, we propose to split it into $L$ subsets. In every iteration, the network is exposed to only one of the training subsets, while other subsets remain idle. Consequently, the network will train, i.e. carry out backpropagation, on a particular subset once every $L$ iteration. Therefore, the neural network will be exposed to the full training dataset after $L$ training epochs. 
This approach differs from conventional training methods, in which the entire training dataset is typically used to update the weights of the neural network in every epoch using the mini-batch gradient descent method. Specifically, in this method, a subset of the dataset (called a mini-batch) is exposed to the network, the error is calculated, and the weights are updated accordingly. However, due to the large size of the graph nodes and the limited memory capacity of the computer, the mini-batch size in our case is limited to 8.
In both the conventional and proposed effective schemes, we use the mini-batch gradient descent method.

However, the difference between the conventional approach \textcolor{black}{(Algorithm \ref{alogorithm2})} and our proposed method \textcolor{black}{(Algorithm \ref{alogorithm1})} is that the mini-batch gradient descent is implemented on a subset, rather than the entire dataset. This allows us to take advantage of the benefits of mini-batches when training a network on large data samples.

\textcolor{black}{\begin{minipage}{0.96\columnwidth}
\begin{algorithm}[H]
\caption{Effective Training Scheme with Data-Dropout Strategy}
\textcolor{black}{\begin{algorithmic}[1]{
\State Let $L$ be the total number of subsets
\State Let $N$ be the total number of epochs
\For{$i = 0$ \textbf{to} $N-1$}
    \State Compute subset index $J = i \mod L$
    \State Extract subset $S_J$ from the training set
    \State Train the model using subset $S_J$
\EndFor
}\end{algorithmic}} \label{alogorithm1}
\end{algorithm}
\end{minipage}
}
\begin{minipage}{0.96\columnwidth}
\textcolor{black}{\begin{algorithm}[H]
\caption{Conventional Training Scheme without Data-Dropout Strateg}
\makeatletter 
\setcounter{ALG@line}{7} 
\makeatother
\textcolor{black}{\begin{algorithmic}[1] 
\State Let $N$ be the total number of epochs
\For{$i = 0$ \textbf{to} $N-1$}
    \State Train the model using the entire training dataset
\EndFor
\end{algorithmic} }\label{alogorithm2}
\end{algorithm}}
\end{minipage}
\vspace{5pt}

\textcolor{black}{To show the effectiveness of the proposed training scheme, it was used to train GTNN and then compared to the conventional training scheme on the same training dataset, namely MOD \cite{M8} dataset. MOD contains sequences of event streams, for benchmarking learning-based motion segmentation models.} Fig. \ref{fig:with_without_dropout} depicts the loss curves obtained in both cases where it can be clearly observed that the proposed effective training scheme has resulted in faster training and better network learning capacity. \textcolor{black}{Although the loss value obtained using conventional training in the first 400 training epochs was less than that obtained using our proposed training scheme, the training time was five times slower as indicated in Table \ref{tab:with_without_dropout_strategy}. After 400 epochs, the loss was still decreasing using the proposed training scheme, which proves the enhanced network learning capacity. 
Upon testing the trained models on unseen data, the results demonstrated the effectiveness of the proposed training scheme where better segmentation accuracy was achieved compared to conventional training, as listed in Table \ref{tab:with_without_dropout_strategy}. }

\begin{figure}[!t]
\centering
 \includegraphics[width=0.490\textwidth]{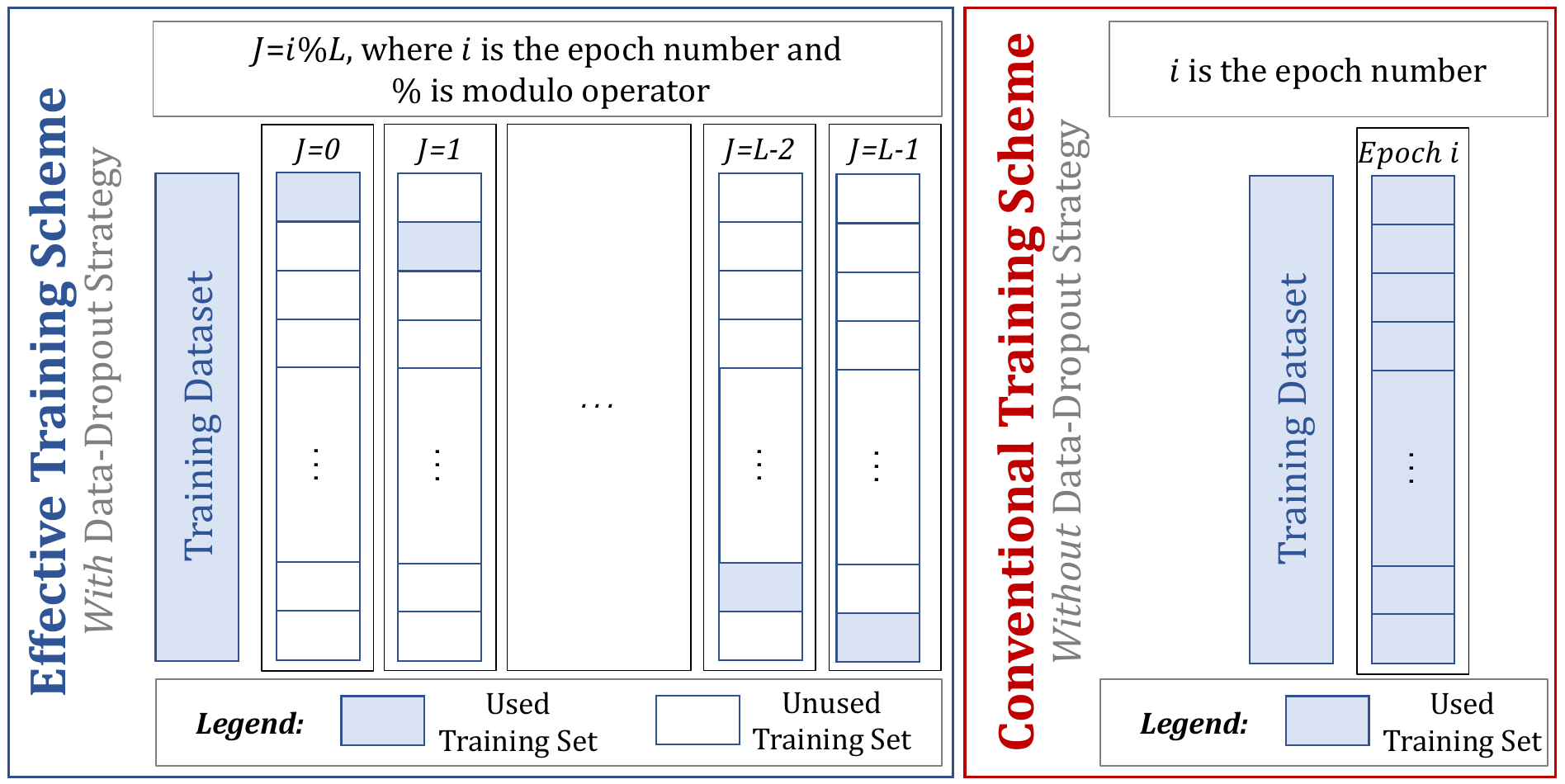}
  \setlength{\belowcaptionskip}{-8pt}   

 \caption{The framework of the proposed effective training scheme compared to the conventional training scheme. Note that $i$ is the current epoch number, $L$ is the number of training subsets, $J=i\%L$ where $\%$ is the modulo operator.}
  \label{fig:effective_regular_training_strategy}
\end{figure}

\begin{figure}[!t]{
\centering
 \includegraphics[width=0.50\textwidth]{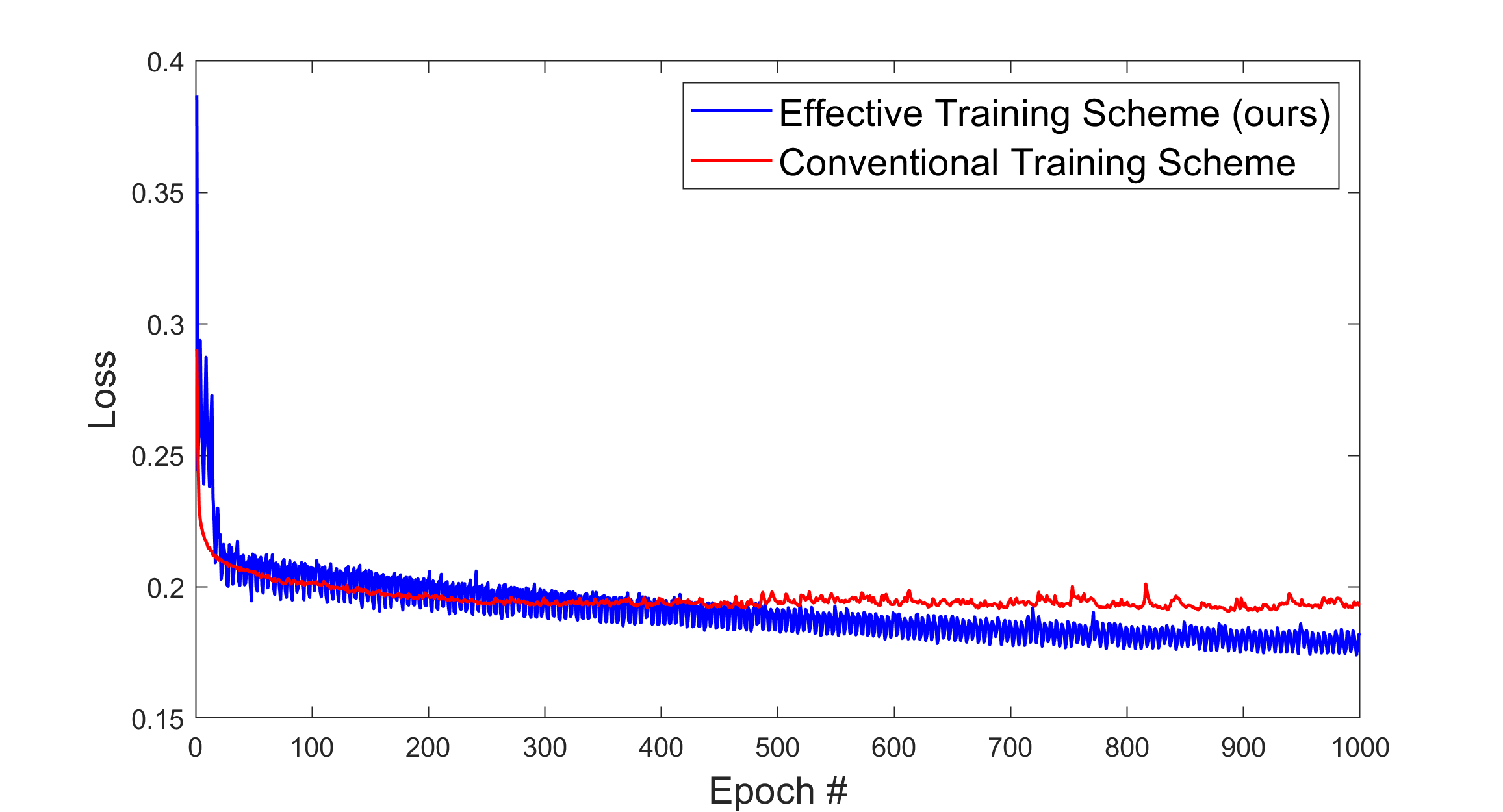}
 
  \setlength{\belowcaptionskip}{-4pt}   

 \caption{Loss curves were obtained while training the network with the proposed effective and conventional training schemes (with and without data dropout strategy, respectively). }
  \label{fig:with_without_dropout}}
\end{figure}
\raggedbottom

\begin{table}[!tb]
\caption{Comparison between the training execution time and testing performance using the proposed effective and conventional training schemes.}
\label{tab:with_without_dropout_strategy}
\begin{adjustbox}{width=0.49\textwidth}
{Large
\begin{tabular}{|c|c|c|}
  \specialrule{.15em}{.1em}{.1em}

\begin{tabular}{c} Execution/Evaluation \\Metrics\end{tabular} &\begin{tabular}{c} Effective Training Scheme \\(with Data-Dropout strategy)\end{tabular} &
\begin{tabular}{c} Conventional Training Scheme \\(without Data-Dropout strategy)\end{tabular}
 \\  \specialrule{.15em}{.1em}{.1em} 
\multicolumn{3}{|c|}{Training} \\ \hline

\begin{tabular}{c} Required time to execute\\  one epoch (s) \end{tabular} & \textbf{42} & 225\\
  \specialrule{.15em}{.1em}{.1em} 

\multicolumn{3}{|c|}{Testing}\\ \hline

($TP$ , $FP$ , $FN$ , $TN$)& (97681, 96816, 58823, 2216680) &(82264, 145343, 74248, 2168062) \\

\textit{F1} score &  \textbf{55.6\%} & 42.8\% 
\\
\textit{Recall} score & \textbf{62.4\%} & 52.5\% 
 \\ \specialrule{.2em}{.1em}{.1em} 
 
\end{tabular}}
\end{adjustbox}
\end{table}
 \raggedbottom

\vspace{20cm} 

\subsection{Dynamic Object Mask-aware Event Labeling (DOMEL)} \label{DOMEL_section}
\textcolor{black}{DOMEL is a new framework capable of generating annotations for event streams from a variety of event camera types, including but not limited to DAVIS346c and DVXplorer. This flexibility is made possible by leveraging reference frames from the DAVIS346c camera, which provides both event streams and RGB images, unlike the DVXplorer that only captures event streams. By accommodating different sensing modalities, DOMEL ensures that event labeling can be consistently applied across diverse camera technologies, facilitating the development and testing of learning-based models for the task at hand. }

To validate the generalization capability of the proposed GTNN motion segmentation algorithm, new experimental sequences are recorded in various domains using event cameras with different resolutions. For a thorough quantitative and qualitative evaluation of GTNN using these sequences, a labeling method must be devised to facilitate comparison of the GTNN prediction against the ground truth.  

To that end, inspired by the success of the KoGTL \cite{AlkendiY}, we propose Dynamic Object Mask-aware Event Labeling (DOMEL), which is an offline approach for annotating event data for motion segmentation applications. Every event in the recorded stream is assigned a label; foreground event or background event. The labeling process requires as input the corresponding gray-scale frame; which can be captured using a frame-based sensor, working simultaneously alongside the event camera. 
Hence, DOMEL allows labeling event streams recorded using event cameras that do not generate grayscale images such as DVXplorer. The frame-based sensor and the event camera need to visualize the same scene, with the same field of view.

In the following sections, the experimental setup for recording the event dataset will be described and the labeling process will be explained in detail. Please refer to Fig. \ref{fig:dvs_proposedApproach1} for an illustration of DOMEL and how it fits in the overall proposed framework.

\subsubsection{Experimental setup}


Two dynamic active pixel vision sensors; DAVIS346C and DVXplorer, are mounted side-by-side on a tripod, to capture a dynamic scene. DAVIS346C and DVXplorer have a spatial resolution of 346$\times$260 and 640$\times$480, a bandwidth of 12 and 165 MEvent/s, and a dynamic range of 120 and 90 dB, respectively. The cameras are moved along various trajectories; i.e. a sequence of translations and rotations, in environments with various scene geometries where objects of various types and sizes are randomly moving. 

Three measurements were recorded; (1) DAVIS346C event streams, (2) DVXplorer event streams, and (3) Grayscale images which capture intensity measurements of the dynamic scene. The grayscale images are obtained from the frame output of the DAVIS346C sensor and are denoted as active pixel sensor (APS) images hereafter. It is worth noting that in absence of the frame output of the event camera, it is possible to use a standard camera to capture the same scene alongside the event \textcolor{black}{camera.}


\subsubsection{Labeling Framework}
DOMEL approach includes four main stages, event-image synchronization, raw event-edge fitting, spatially shifted event-mask fitting, and event labeling, as illustrated in Fig. \ref{fig:domel}.  
\begin{figure*}[]
\centering
 \includegraphics[width=0.85\textwidth]{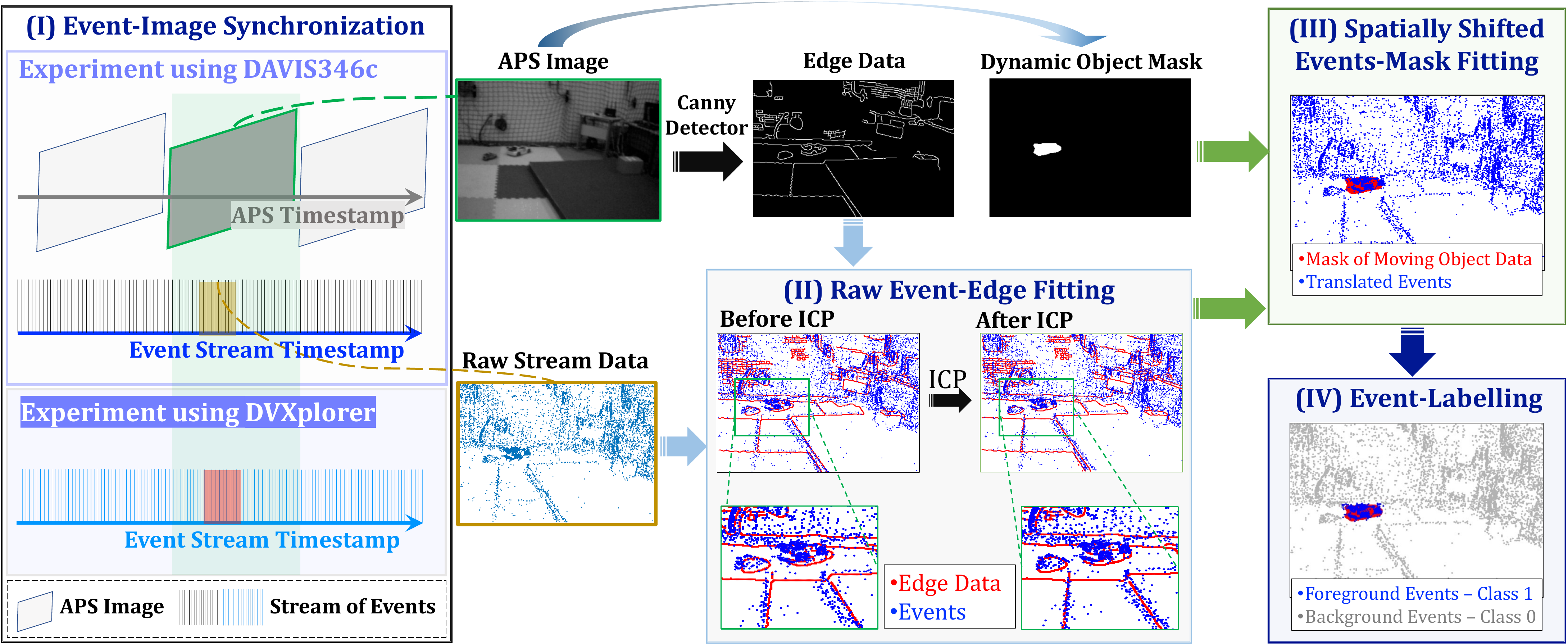}
 \setlength{\belowcaptionskip}{-8pt}   

  \caption{DOMEL framework for Event-based Motion Segmentation (EMS). DOMEL is a novel event labeling methodology developed to classify events, acquired when the camera is in motion, into two main classes: foreground or background events. The proposed DOMEL works irrespective of the sensor resolution, and hence any event camera may be used to record the event streams. A frame-based sensor is needed to capture intensity images corresponding to the recorded events, which will assist in event labeling.}
  \label{fig:domel}
\end{figure*}

\noindent\paragraph{Event-Image Synchronization}
Synchronization of the recorded event streams from DAVIS346c and DVXplorer with the corresponding APS images is vital to the success of the proposed labeling technique, since events are matched to the corresponding APS frames captured at the same time. 
Synchronization is achieved by recording sensor data through a single ROS\cite{ros} node, and hence a common clock is used to record the measurements' timestamps, which can be seamlessly matched to obtain synchronized measurement pairs, as shown in Fig. \ref{fig:domel}-(I). 


\noindent\paragraph{Raw Event-Edge Fitting}
The iterative closest point (ICP) fitting technique \cite{icp} is used to fit event streams to their corresponding APS canny \cite{canny1986computational} edge data. Due to the high temporal resolution of event data acquisition, fitting of events to edges is performed in several iterations. 
If the scene exhibits high dynamicity or the motion of the camera is fast, event streams are generated at a higher resolution than APS frames. Subsequently, events, particularly those in between two APS frames, would slightly deviate from the APS edges. To compensate for this deviation, ICP is used to perform a spatial shift (i.e., rotation and translation) to the events, which will facilitate matching them to APS edges, as presented in Fig. \ref{fig:domel}-(II). The spatially shifted events will be used for further processing in the next stages. 


\noindent\paragraph{Spatially Shifted Events-Mask Fitting}
APS Frames capturing dynamic objects in the scene are first processed to generate masks of the objects. These masks could be obtained using any object masking algorithm, such as image segmenter \cite{image_segmenter}. The spatially shifted raw events are fitted to the corresponding masked-object frames using ICP, similar to stage (II), in several iterations for both cameras, as shown in Fig. \ref{fig:domel}-III. This is due to the high temporal resolution of the sensors, especially when capturing a dynamic scene, and the camera is in motion.  

\noindent\paragraph{Event-Labeling}
In the last stage, events that are fitted to the corresponding masked-object frame are labeled as foreground events representing the moving objects captured when the camera is in motion (Class 1), as shown in Fig. \ref{fig:domel}-(IV). Whereas events that are not part of the masked pixels of the frame are considered background events (Class 0).

\textcolor{black}{Our framework is efficient due to the use of ICP fitting which accommodates the high temporal resolution of DVS data acquisition. Events could slightly misalign with the edges in images because of slight timing mismatches between the time when events and image frames are captured. Therefore, ICP is used to precisely align the edges of event data and image pixels, correcting any slight shifts in their pixel positions. This ensures that our annotations accurately overlay the dynamic object mask, which is a significant improvement compared to other annotation schemes (\cite{m6_evimo, EVIMO2} that might not fix these small but important differences.}

\subsubsection{\textcolor{black}{EMS-DOMEL Dataset}}
\textcolor{black}{
In this section, the collection and annotation of the Event-based Motion Segmentation dataset (EMS-DOMEL) using the DOMEL framework is discussed. The dataset captures multiple independently moving objects in an indoor environment using a moving event camera. The sequences capture a variety of scenes, multiple objects moving at various speeds and in random paths, and unknown camera motions with various sensor resolutions. 
Table \ref{tab:DOMEL} presents a summary of event sequences captured using two cameras with different resolutions, detailing the number of detected objects, dynamic object events, and background events for each sequence. For instance, in DOMEL-Seq01 captured using a DAVIS346c camera, two moving objects were detected, resulting in 42,980 dynamic object events and 440,650 background events. Similarly, DOMEL-Seq07 to DOMEL-Seq09 which were captured using the DVXplorer camera, showcase varying numbers of moving objects resulting in different amounts of dynamic and background events as listed in the table. 
The ground truth labels facilitate rigorous assessment through metrics such as F1 score, Recall, Intersection over Union (IoU), and Detection Rate, which serve as evaluation benchmarks detailed in the following section (Section \ref{sec:evaluation_metrics}). 
This dataset is essential for advancing learning-based motion segmentation models, offering a wealth of data for both training algorithms and their corresponding performance evaluation metrics.}

\begin{table}[htbp]
  \centering
  \caption{Summary of EMS-DOMEL Dataset.}
  \label{tab:DOMEL}
  \resizebox{0.5\textwidth}{!}{\large%
  \begin{tabular}{@{}ccccc@{}} 
    \toprule
    \textbf{EMS-DOMEL} & \textbf{Sensor Type} & \textbf{\# Dynamic Objects} & \textbf{\# Dynamic Object Events} & \textbf{\# Background Events} \\
    \midrule
    Seq01 & DAVIS346c & 2  & 42,980 & 440,650 \\
    Seq02 & DAVIS346c & 2 & 52,176 & 494,378 \\
    Seq03 & DAVIS346c & 4  & 282,397 & 640,531 \\
    Seq04 & DAVIS346c & 2  & 37,220 & 146,949 \\
    Seq05 & DAVIS346c & 2  & 112,980 & 816,152 \\
    Seq06 & DAVIS346c & 2  & 66,989 & 560,096 \\
    Seq07 & DVXplorer & 3  & 616,467 & 2,923,756 \\
    Seq08 & DVXplorer & 3  & 889,215 & 5,953,718 \\
    Seq09 & DVXplorer & 1  & 77,314 & 4,435,217 \\
    \bottomrule
  \end{tabular}%
  }
\end{table}

}
\section{Experimental Evaluations}\label{sec:results}

\subsection{Training and Testing Datasets}
The proposed GTNN will be trained and evaluated using publicly available datasets including MOD, EV-IMO, and EV-IMO2 datatsets. 
MOD \cite{M8} is a simulated dataset targeted for learning-based motion segmentation models for event cameras. The environment captured in this dataset is a highly textured synthetic indoor room, where one to three dynamic objects appear intermittently, while the camera is moving. 
EV-IMO \cite{m6_evimo} dataset, on the other hand, contains various event sequences recorded in a real lab environment. The sequences exhibit varying levels of complexity in terms of the motion of the camera, dynamic objects' number, motion and speed randomness, and occlusion, feature-rich background texture, and a range of lighting conditions in the lab. Five different sequences of the EV-IMO dataset including {Boxes}, {Floor}, {Wall}, {Table}, and {Fast} are used. 
Both MOD and EV-IMO datasets were recorded using the same event cameras, namely DAVIS 346c which generates events within a spatial resolution of 346 $\times$ 260 pixels.

EV-IMO2 \cite{EVIMO2} is a second version of EV-IMO, recorded using a higher resolution event camera (480$\times$640 pixels). EV-IMO2 contains sequences of small-to-large and slow-to-fast moving objects with various illumination conditions. It is targeted for motion segmentation approaches, depth estimation, optical flow, visual odometry, and SLAM methods \cite{EVIMO2}. 
Finally, we release as part of this paper a new event dataset, called EMS-DOMEL. This dataset is recorded locally in our lab facilities and includes several scenes of one to four dynamic objects captured using a moving camera. Event sequences are labeled using the DOMEL approach, described in and labeled using the proposed Section \ref{DOMEL_section}. EMS-DOMEL will be used exclusively for testing. None of the sequences will be exposed to GTNN during the training phase and hence, the results will demonstrate the model's ability to generalize across various environments. 

To train the GTNN algorithm, supervised learning is performed using the backpropagation technique. Pytorch is used to construct all the neural networks for training and testing. The Adam optimizer with a learning rate of 0.001 is used to execute training process to minimize the Dual Focal Loss (DFL). DFL was proposed in \cite{DFL} to tackle the challenge of imbalanced training datasets for neural classification.

\subsection{Evaluation Metrics}
\label{sec:evaluation_metrics}
To quantitatively evaluate the performance of our proposed approach, four evaluation metrics are used including \textit{F1} score, \textit{Recall} metric, \textit{Intersection over Union} (\textit{IoU}), and \textit{Detection Rate (DR)}. The first two metrics are commonly used for object detection algorithms \cite{object_detection_metric22}. Particularly, \textit{Recall} measures the ratio of the correctly detected dynamic objects to the true number of dynamic objects in the scene, as defined in \textcolor{black}{Eq. (\ref{recall})}. Whereas \textit{Precision} measures the the ratio of the correctly detected dynamic objects to the total positive detections, as defined in \textcolor{black}{Eq.(\ref{precision})}. \textit{F1} score, in consequence, computes the harmonic mean of \textit{Precision} and \textit{Recall}, as defined in \textcolor{black}{Eq. (\ref{f1})}.

\begin{equation}\label{recall}
   { \textit{Recall}=\frac{TP}{TP + FN}\textcolor{black}{,}}
\end{equation}

\begin{equation}\label{precision}
   { \textit{Precision}=\frac{TP}{TP + FP}\textcolor{black}{,}}
\end{equation}

\begin{equation}\label{f1}
 {   \textit{F1}=2* \frac{\textit{Recall} * \textit{Precision} }{\textit{Recall} + \textit{Precision}}\textcolor{black}{,}}
\end{equation}

\noindent where $TP$, $FP$, $TN$, and $FN$ are the number of true positives,  false positives, true negatives, and false negatives pixels, respectively. $TP$ and $FP$ indicate the number of events that are correctly and incorrectly predicted as dynamics objects-related events, respectively. While $TN$ and $FN$ indicate the number of events that are correctly and incorrectly predicted as background-related events.
\textit{Recall} and \textit{F1} are used to assess the capability of the proposed algorithm to distinguish between foreground and background events during the ablation study as discussed in the supplementary material.

\textit{Detection Rate (DR)} and \textit{Intersection over Union (\textit{IoU})}, are two standard metrics used to quantitatively evaluate the performance of SOTA motion segmentation models as reported in \cite{SpikeMS, M9_s128_Mitrokhin_2018, M8}. \textit{DR} is the area overlap between the bounding box containing the predicted foreground events and the corresponding ground truth, as introduced in \cite{M9_s128_Mitrokhin_2018, M8}. The detection is considered successful if the following criteria is met:

\begin{equation}\label{eq:DR}
{\text{Success if:      }   (D \cap G) > 0.5    \text{    and    }    (D \cap G) > (D \cap G^c)\textcolor{black}{,}}
\end{equation}

\noindent where $D$ is the predicted bounding box (or convex hull), $G$ is the corresponding ground truth bounding box (or convex hull), and $G^c$ denotes the complement of the $G$ set. 

\textit{Intersection over Union (\textit{IoU})} is the most commonly used evaluation metric for segmentation algorithms and has been used for event-based algorithms as reported in \cite{SpikeMS}. \textit{IoU} is defined in the following equation: 
\begin{equation}\label{eq:iou}
{\textit{IoU} = \frac{S_D \cap S_G}{S_D \cup S_G}\textcolor{black}{,}}
\end{equation}

\noindent where $S_D$ is refers to the predicted object(s) segmentation mask and $S_G$ is the corresponding ground truth mask. To generate a dense segmentation mask, $S_D$ and $S_G$, the events are first projected into the 2D frame captured within the specified time window. 
Then, the convex hull outlining the events that belong to the dynamic object is computed. Consequently, all pixels within that convex hull are considered to be foreground events. 
\vspace{-0.2cm}
\subsection{Quantitative Evaluation}\label{sec:qaun}
In this section, we first provide a comparison between the performance of our GTNN model against SOTA learning-based motion segmentation model, SpikeMS as proposed in \cite{SpikeMS}, in terms of \textit{IoU}\%.

Motion segmentation is carried out on the evaluation sequences from EV-IMO and MOD datasets, which were not exposed to the network during training. The results obtained using GTNN and SpikeMS \cite{SpikeMS} are reported in Table \ref{tab:Iou_spike}. In addition, the \textit{IoU}\%  metric obtained using GTNN for EV-IMO2 evaluation dataset is 49.76\%. The segmentation results, based on the \textit{IoU}\% metric, obtained using GTNN outperform SpikeMS \cite{SpikeMS} in almost all sequences with an average of 9.4\%. This is clearly due to the fact that graph transformers with the attention mechanism reveal and exploit the spatiotemporal correlations between the events that belong to the moving object and segregate them from background events. The network structure also aids extraction of global feature information and aggregates them with the local features extracted by the graph transformer's encoder-decoder units. Our network is able to distinguish between foreground and background events, especially when the relative speed of the camera and the moving objects is distinct. 

\begin{table*}[!tb]
\caption{Performance of the GTNN algorithm compared to SOTA motion segmentation learning-based method, SpikeMS \cite{SpikeMS} on EV-IMO and MOD event-based datasets. Segmentation results are compared using the \textit{IoU} (in \%) $\uparrow$ metric.}
\label{tab:Iou_spike}
\begin{adjustbox}{width=0.99\textwidth}
\setlength{\arrayrulewidth}{0.5pt}
{ \Huge 
\begin{tabular}{|c|ccc|ccc|ccc|ccc|ccc|ccc|}

  \specialrule{.15em}{.1em}{.1em} 
                                                & \multicolumn{15}{c|}{EV-IMO dataset}                                                                                                                                                                                                                                                                                                                                                                                                                           & \multicolumn{3}{c|}{}                                                                       \\
                                                  & \multicolumn{3}{c|}{Boxes}                                                                  & \multicolumn{3}{c|}{Floor}                                                             & \multicolumn{3}{c|}{Wall}                                                             & \multicolumn{3}{c|}{Table}                                                             & \multicolumn{3}{c|}{Fast}                                                                  & \multicolumn{3}{c|}{\multirow{-2}{*}{MOD dataset}}                                          \\
\multirow{-3}{*}{Method}                          & 100 ms                       & 20 ms                        & 10ms                         & 100 ms                      & 20 ms                        & 10ms                     & 100 ms                      & 20 ms                       & 10ms                     & 100 ms                       & 20 ms                       & 10 ms                    & 100 ms                       & 20 ms                        & 10 ms                        & 100 ms                       & 20 ms                        & 10ms                         \\
\specialrule{.2em}{.1em}{.1em} 

SpikeMS\textsuperscript{\textdagger} \cite{SpikeMS}                                 & \textbf{{  61±7}}  & \textbf{{  65±8}}  & {  -}     & {  60±5} & {  53±16} & {  -} & {  65±7} & {  63±6} & {  -} & {  52±13} & {  50±8} & {  -} & {  45±11} & {  38±10} & {  -}     & {  68±7}  & {  65±5}  & {  -}     \\
{  \textbf{GTNN model (ours)}} & { 49±36} & { 40±31} & { 34±30} & \textbf{77±13 }                      & \textbf{68±19}                        & 63±22                    & \textbf{77±15 }                      & \textbf{65±21}                       & 58±23                    & \textbf{71±23 }                       & \textbf{62±27 }                      & 56±29                    & \textbf{{ 65±22}} & \textbf{{ 49±25}} & { 43±25} & { \textbf{78±21}} & { \textbf{73±30}} & { 73±31}\\  \specialrule{.15em}{.1em}{.1em} 

\multicolumn{19}{c}{\textsuperscript{\textdagger}{Results taken directly from \cite{SpikeMS}}}

\end{tabular}}
\end{adjustbox}
\end{table*}

Additionally, the performance of the proposed GTNN model is compared to classical hand-crafted motion segmentation methods \cite{M9_s128_Mitrokhin_2018,M8} and SpikeMS \cite{SpikeMS} based on the detection rate (\textit{DR\%}) metric, as reported in Table \ref{tab:DR_classical_spike}. Note that our GTNN model performs event-based motion segmentation without any prior knowledge about the scene geometry, the number of dynamic objects within the scene, and without any initialization stage, as opposed to the SOTA classical models \cite{M9_s128_Mitrokhin_2018,M8}. The results of the SOTA models are taken directly from their corresponding publications \cite{M9_s128_Mitrokhin_2018,M8,SpikeMS}. Our approach outperforms SpikeMS \cite{SpikeMS}, the learning-based model, and has comparable performance against offline classical-based approaches \cite{M9_s128_Mitrokhin_2018,M8}.  
It can be concluded that the effectiveness of the proposed GTNN algorithm is validated across multiple unseen event streams acquired by a moving camera in a dynamic environment. The proposed algorithm could be integrated with other vision-based modules to ensure safe robotic navigation in an unknown and dynamic environment under challenging illumination conditions. 

\begin{table}[]
\caption{Comparison with SOTA \textit{learning-based} and \textit{classical} approaches on sequences from EV-IMO and MOD event-based datasets, in terms of \textit{Detection Rate (DR)} (in \%) metric. Note that "$\textit{C}$" and "\textit{L}" indicate the type of the adopted approach; classical or learning, respectively.}
\label{tab:DR_classical_spike}
\begin{adjustbox}{width=0.49\textwidth}
{ \Huge 
\begin{tabular}{|c|c|>{\centering\arraybackslash}p{2.2cm}|>{\centering\arraybackslash}p{2.2cm}|>{\centering\arraybackslash}p{2.2cm}|>{\centering\arraybackslash}p{2.2cm}|>{\centering\arraybackslash}p{2.2cm}|>{\centering\arraybackslash}p{2.2cm}|}
  \specialrule{.15em}{.1em}{.1em} 
                                               &         & \multicolumn{3}{c|}{EV-IMO dataset}          & \multicolumn{3}{c|}{MOD dataset}             \\
\multirow{-2}{*}{Method}                &    \multirow{-2}{*}{Type}            & 100ms         & 20ms        & 10ms        & 100ms         & 20ms        & 10ms        \\ 
                          \specialrule{.15em}{.1em}{.1em} 
  
{Mitrokhin et al.\textsuperscript{\textdagger} \cite{M9_s128_Mitrokhin_2018}}           &\textit{C}   & \multicolumn{3}{c|}{{48.79}} & \multicolumn{3}{c|}{{70.12}} \\ 

  \specialrule{.05em}{.05em}{.05em}

{  0-MMS\textsuperscript{\textdaggerdbl}  \cite{M8} } &\textit{C}  & \multicolumn{3}{c|}{81.06}                   & \multicolumn{3}{c|}{82.35}                   \\\hline

SpikeMS.\textsuperscript{\textdagger} \cite{SpikeMS}                                &   \textit{L} & \multicolumn{3}{c|}{65.14}            & \multicolumn{3}{c|}{68.82}  \\
  \specialrule{.05em}{.05em}{.05em} 

{  \textbf{Ours (GTNN model)}}    &  \textit{L}  & 73             & 56           & 48          & 84             & 77           & 75    \\   \specialrule{.15em}{.1em}{.1em} 
\end{tabular}}
\end{adjustbox}
\end{table}
\raggedbottom

\subsection{Qualitative Evaluation}
\label{seq:qualitative}

In this section, the testing sequences from the publicly-available datasets, MOD, EV-IMO and {EV-IMO2}, and our recorded experiments are used to qualitatively evaluate the motion segmentation performance. Note that, the tested sequences are not exposed to our network during training and the output of GTNN has the same structure as the 3D graph input. However, for visualization purposes, the results are projected into 2D frames. The results obtained from the proposed model are projected against the approximate ground truth and SOTA models SpikeMS \cite{SpikeMS} and Zhou et al. \cite{M10_Zhou21tnnls}.

\subsubsection{EV-IMO dataset}
\paragraph{GTNN vs. learning-based approach (SpikeMS)}

\begin{figure}[!t]
\centering
\setlength{\fboxrule}{0.8pt}%

\centering
\begin{adjustbox}{width=0.47\textwidth}
  \centering
  {\Large
\begin{tabular}{cccc}

\Huge{\begin{tabular}{c} \fontsize{30}{30}\selectfont{APS frame} \\ \fontsize{30}{30}\selectfont{for  visualization only}\end{tabular}} &\fontsize{30}{30}\selectfont{Ground Truth Events}& \fontsize{30}{30}\selectfont{\textbf{GTNN} (\textbf{ours})} &\fontsize{30}{30}\selectfont{SpikeMS\cite{SpikeMS}}
\\


 \fbox{\includegraphics[width=\columnwidth,height=7.1cm]{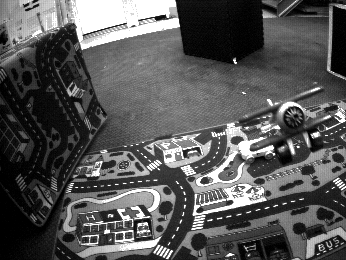}}\centering
&\fbox{\includegraphics[width=\columnwidth,trim={3.8cm 2.4cm 2.8cm 1.5cm},clip]{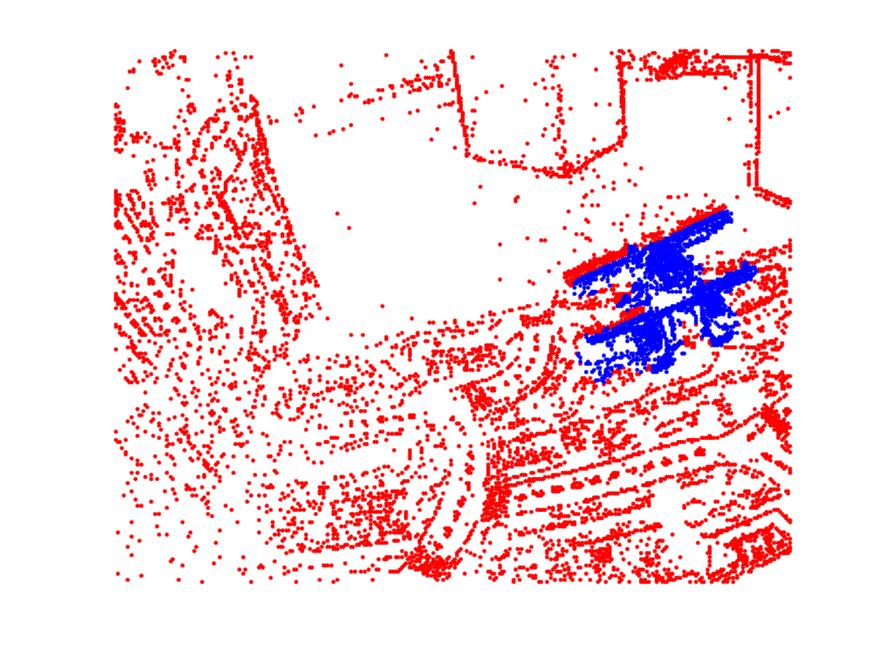}}\centering
&\fbox{\includegraphics[width=\columnwidth,trim={3.8cm 2.4cm 2.8cm 1.5cm},clip]{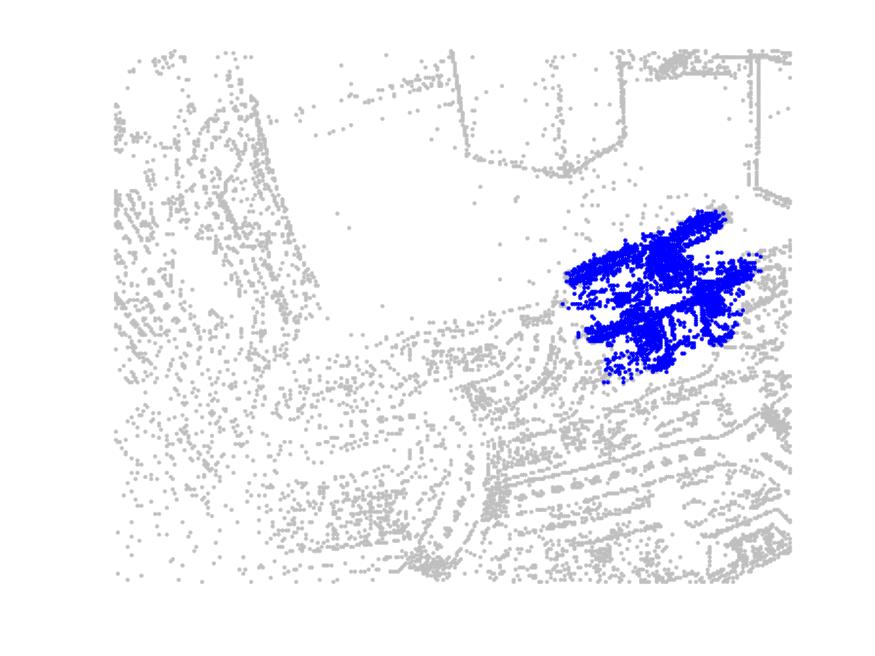}}\centering
&\fbox{\includegraphics[width=\columnwidth,trim={3.8cm 2.4cm 2.8cm 1.5cm},clip]{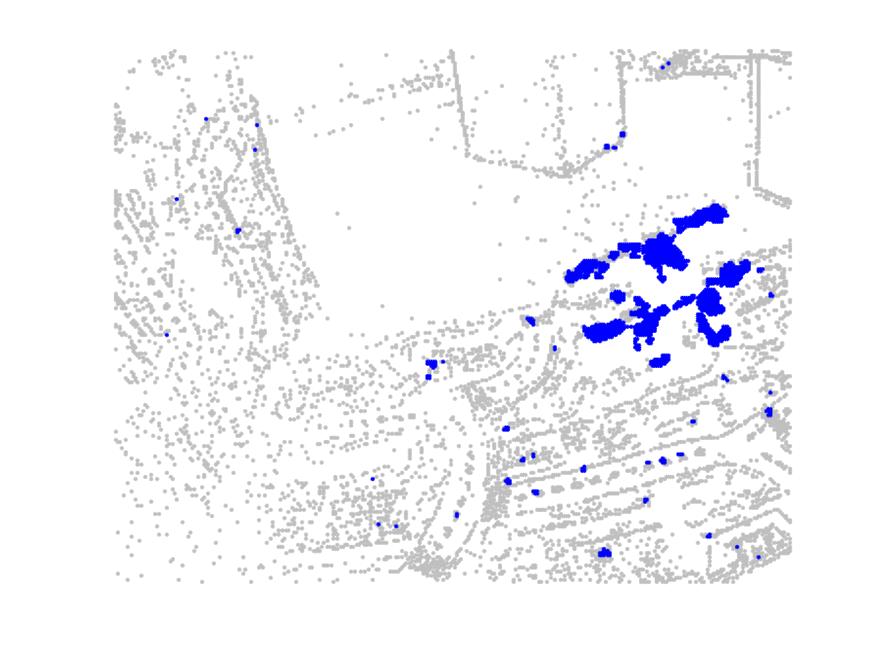}}
\\\\
\multicolumn{4}{c}{\centering\fontsize{30}{30}\selectfont { (A) EV-IMO - Floor }}\\\\

 \fbox{\includegraphics[width=\columnwidth,height=7.1cm]{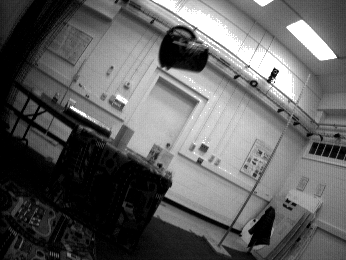}}\centering
&\fbox{\includegraphics[width=\columnwidth,trim={3.8cm 2.4cm 2.8cm 1.5cm},clip]{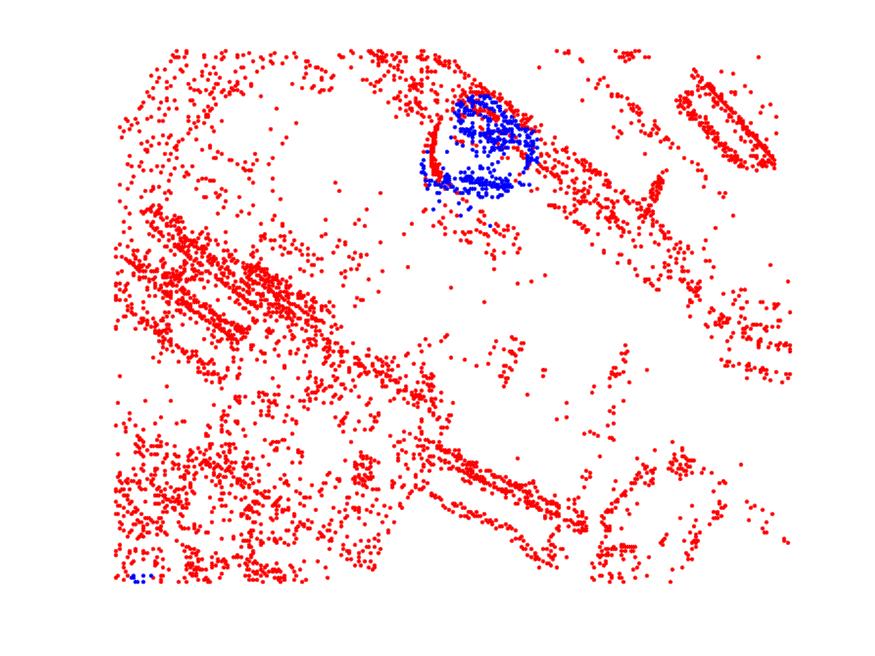}}\centering
&\fbox{\includegraphics[width=\columnwidth,trim={3.8cm 2.4cm 2.8cm 1.5cm},clip]{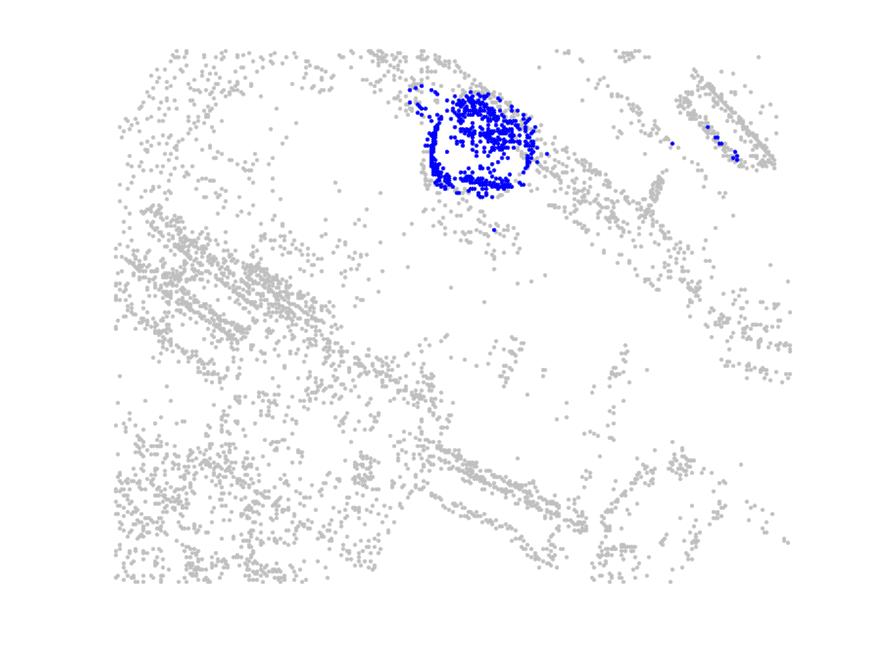}}\centering
&\fbox{\includegraphics[width=\columnwidth,trim={3.8cm 2.4cm 2.8cm 1.5cm},clip]{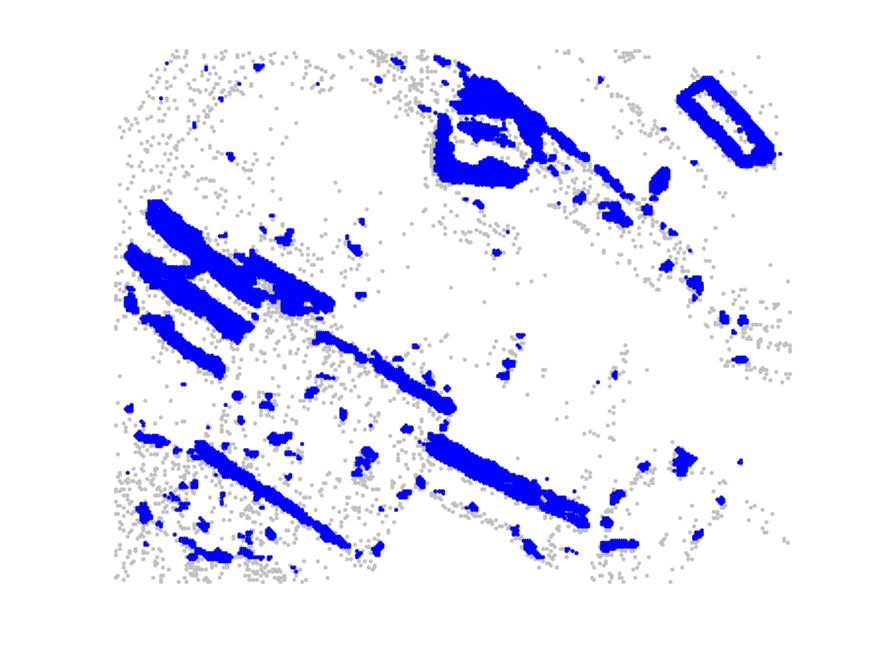}}


\\
\\
\multicolumn{4}{c}{\centering\fontsize{30}{30}\selectfont { (B) EV-IMO - Fast }}\\
\\



 \fbox{\includegraphics[width=\columnwidth, height=7.1cm]{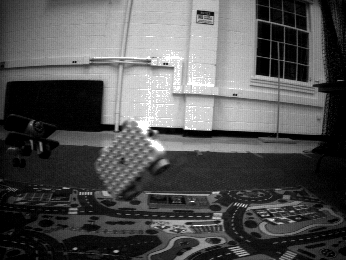}}\centering
&\fbox{\includegraphics[width=\columnwidth,trim={3.8cm 2.4cm 2.8cm 1.5cm},clip]{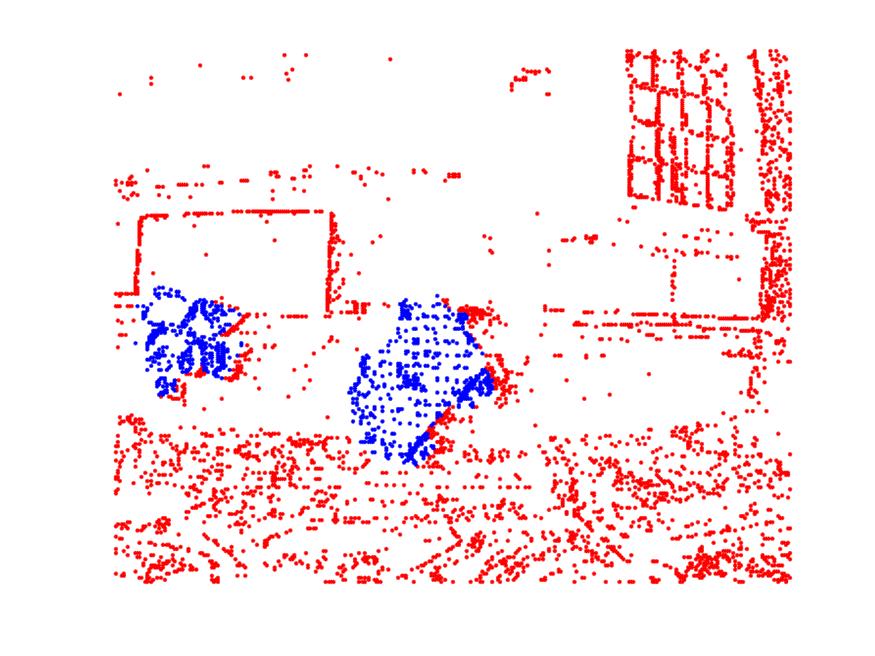}}\centering
&\fbox{\includegraphics[width=\columnwidth,trim={3.8cm 2.4cm 2.8cm 1.5cm},clip]{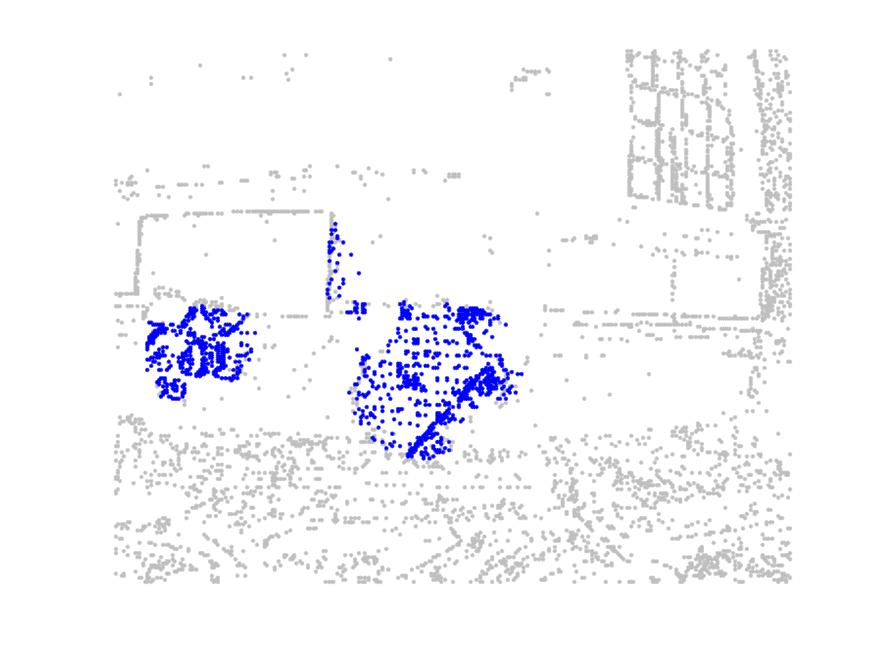}}\centering
&\fbox{\includegraphics[width=\columnwidth,trim={3.8cm 2.4cm 2.8cm 1.5cm},clip]{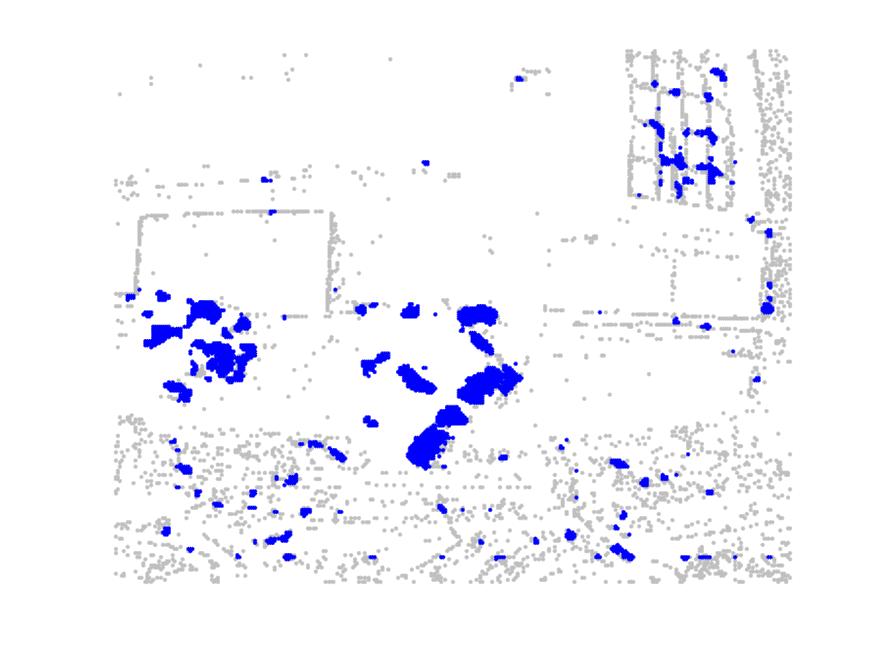}}

\\

\\ 
\multicolumn{4}{c}{\centering\fontsize{30}{30}\selectfont { (C) EV-IMO - Wall}}\\

\end{tabular}}
\end{adjustbox}
 \setlength{\belowcaptionskip}{-12pt}   

\caption{\textcolor{black}{Segmentation results compared with the SOTA learning-based motion segmentation model (SpikeMS \cite{SpikeMS}) on testing sequences from EV-IMO dataset.  }}\label{qual_1}

\end{figure}

In this section, the performance of GTNN will be qualitatively compared to SpikeMS on the EV-IMO dataset. 
Fig. \ref{qual_1} shows selected scenes from three different sequences; (1) Floor, (2) Fast, and (3) Wall, from the EV-IMO dataset. Every row shows the 2D frame, the corresponding ground truth events, GTNN segmentation results, and SpikeMS segmentation results for the selected scenes, respectively. 
It is worth noting that SpikeMS \cite{SpikeMS} segmentation results are obtained using their publicly available pretrained model. For a fair comparison, a 10ms time window is selected to slice and prepare the testing samples for network prediction, since it is the time window selected to train SpikeMS \cite{SpikeMS}.
    
Fig. \ref{qual_1}-(A) depicts a sample scene from the Floor scenario, where a toy plane is moving above a highly textured carpet. The foreground events detected by GTNN solely belong to the moving object, while SpikeMS falsely segmented some of the background events as foreground events. GTNN was able to accurately segment the moving object in presence of significant background variations and motion dynamics, compared to SpikeMS. 

Fig. \ref{qual_1}-(B) shows a another sample scene captured in the {Fast} scenario, where the camera is experiencing harsh motion dynamics (rotation and translation at a high speed) while having a dynamic object within the scene. Our proposed model was able to distinguish events that correspond to dynamic objects in the scene from the background. Conversely, SpikeMS \cite{SpikeMS} has falsely segmented scattered background events, as foreground events.

Furthermore, in Fig. \ref{qual_1}-(C), the performance of the GTNN model and SpikeMS \cite{SpikeMS} are tested on the {Wall} scenario, where multiple moving objects with different sizes, are moving along different random trajectories. Our segmentation results show better interpretation and fine labeling of the dynamic objects compared to SpikeMS \cite{SpikeMS}. This proves the robustness and generalization capability of our model to different environments with significant background textures and motion dynamics. 
Accurate motion segmentation results are important for various applications including event-based instance segmentation \cite{Instance_Segmentation22}, recognition \cite{event_object_classification22}, and localization \cite{hidalgo2022eventOdometry,alkendi2021state} modules. 

\paragraph {GTNN vs. classical-based approach}
To further verify the validity of our proposed model, we analyze the performance of GTNN on other unseen testing scenarios of EV-IMO dataset, particularly from the Boxes and Table sequences, and show qualitatively the results compared to Zhou et al. \cite{M10_Zhou21tnnls} model. Note that Zhou et al. \cite{M10_Zhou21tnnls} model is a recent classical-based approach where motion segmentation is solved as an optimization problem of classical multi-model fitting schemes.

\begin{figure}[!t]
\centering
\setlength{\fboxrule}{0.8pt}%

\centering
\begin{adjustbox}{width=0.47\textwidth}
  \centering
  {\Large
\begin{tabular}{cccc}

\Huge{\begin{tabular}{c} \fontsize{30}{30}\selectfont{APS frame} \\ \fontsize{30}{30}\selectfont{for  visualization only}\end{tabular}} &\fontsize{30}{30}\selectfont{Ground Truth Events}& \fontsize{30}{30}\selectfont{\textbf{GTNN} (\textbf{ours})} &\fontsize{30}{30}\selectfont{Zhou et al. \cite{M10_Zhou21tnnls}}




\\
 \fbox{\includegraphics[width=\columnwidth, height=7.05cm]{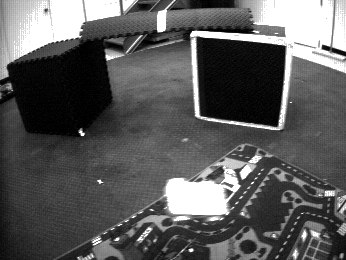}}\centering

&
\fbox{\includegraphics[width=\columnwidth,angle =0,trim={3.8cm 2.4cm 2.8cm 1.5cm},clip]{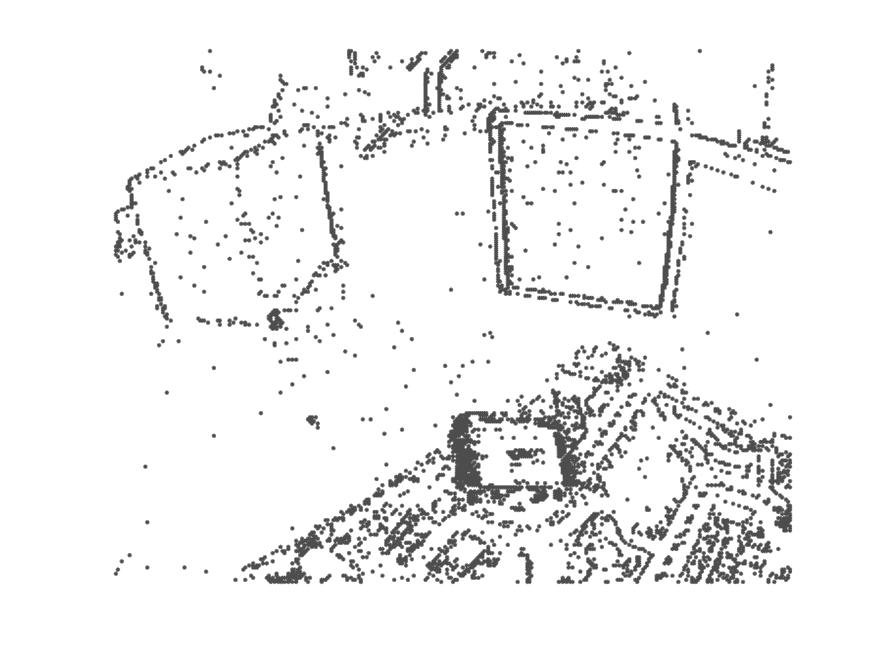}}& 
\fbox{\includegraphics[width=\columnwidth,angle =0,trim={3.8cm 2.4cm 2.8cm 1.5cm},clip]{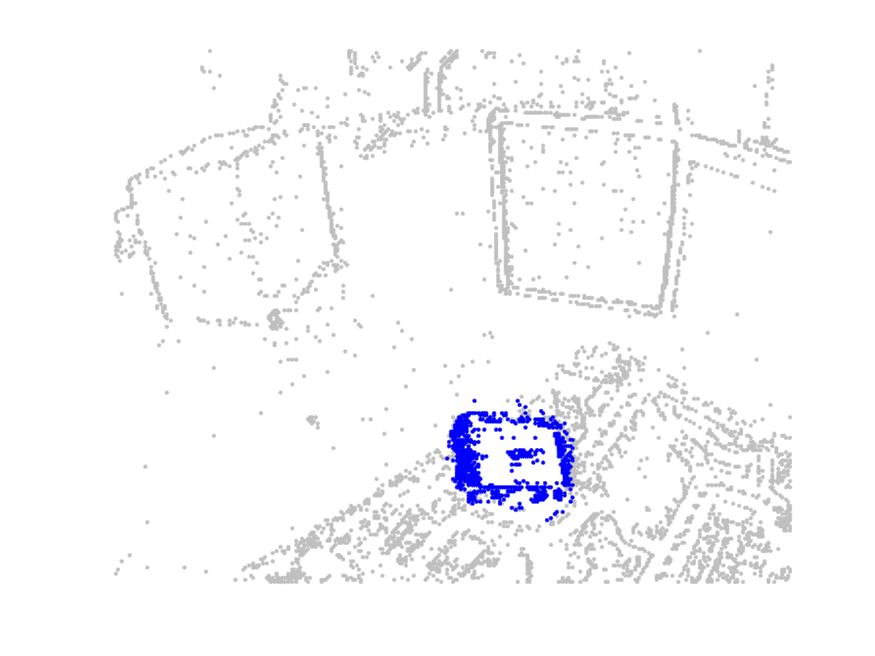}}&    

\fbox{\includegraphics[width=\columnwidth,height=7.05cm,angle =0,clip]{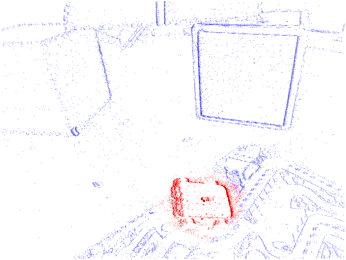}}   
\\ 



 \fbox{\includegraphics[width=\columnwidth, height=7.05cm]{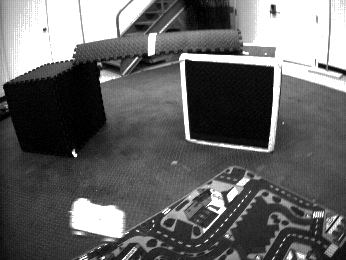}}\centering

&
\fbox{\includegraphics[width=\columnwidth,angle =0,trim={3.8cm 2.4cm 2.8cm 1.5cm},clip]{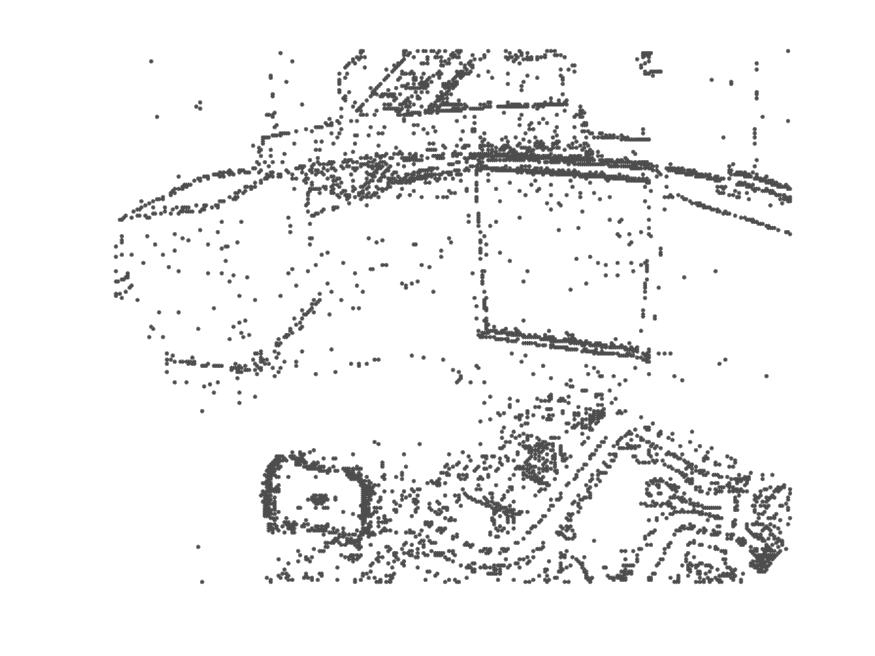}}& 
\fbox{\includegraphics[width=\columnwidth,angle =0,trim={3.8cm 2.4cm 2.8cm 1.5cm},clip]{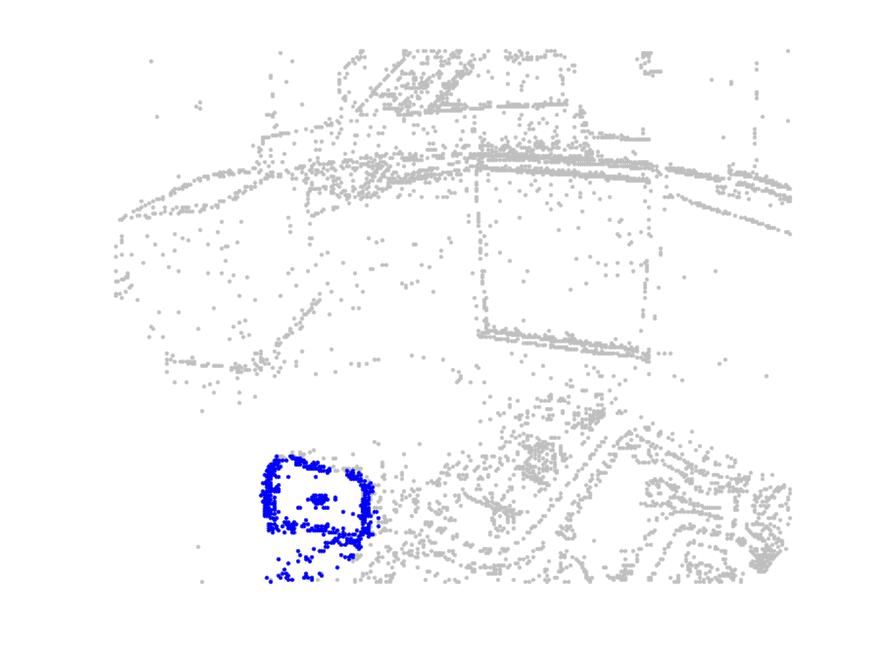}}&    

\fbox{\includegraphics[width=\columnwidth,height=7.05cm,angle =0,clip]{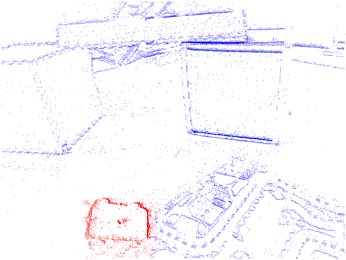}}    

\\
\\
\multicolumn{4}{c}{\centering\fontsize{30}{30}\selectfont {(A) EV-IMO - Boxes seq00 - single dynamic object within the scene}}

\\\\

\fbox{ \includegraphics[width=\columnwidth, height=7.05cm]{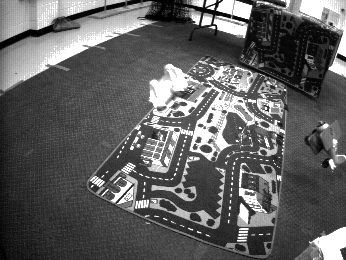}}\centering&
\fbox{\includegraphics[width=\columnwidth,angle =0,trim={3.8cm 2.4cm 2.8cm 1.5cm},clip]{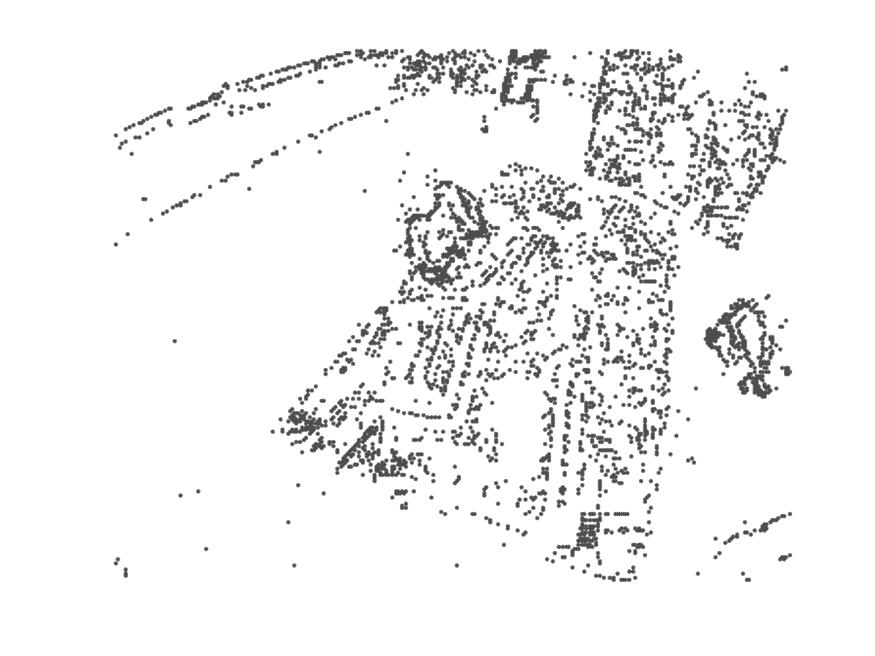}}&
\fbox{\includegraphics[width=\columnwidth,angle =0,trim={3.8cm 2.4cm 2.8cm 1.5cm},clip]{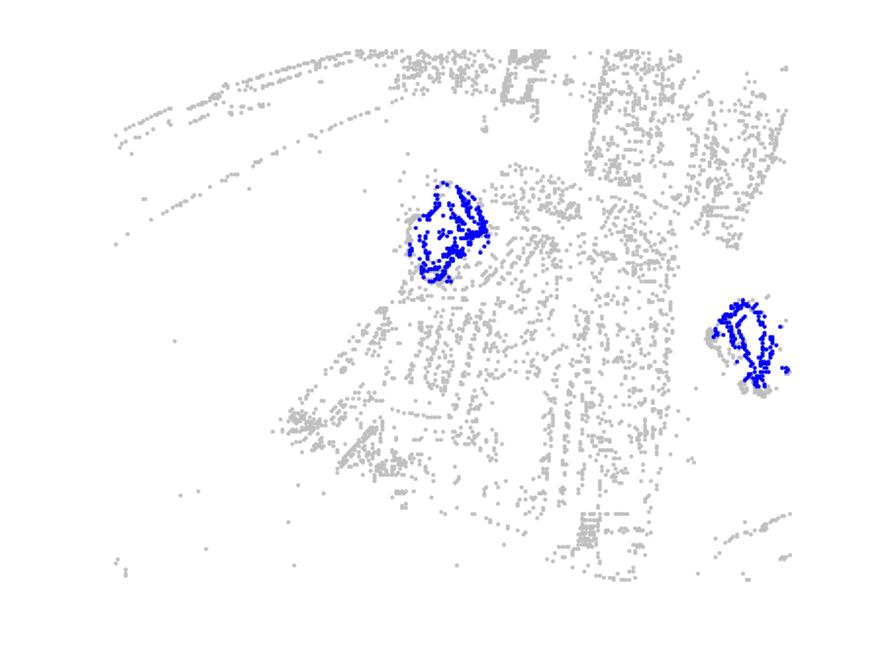}}&
\fbox{\includegraphics[width=\columnwidth,height=7.0cm,angle =0,clip]{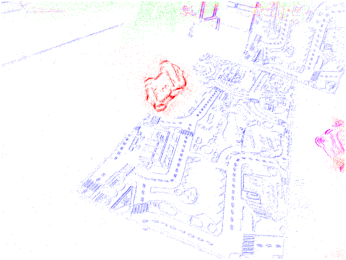}}




\\
 \fbox{\includegraphics[width=\columnwidth, height=7.05cm]{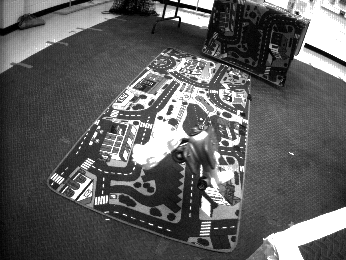}}\centering
&  \fbox{\includegraphics[width=\columnwidth,angle =0,trim={3.8cm 2.4cm 2.8cm 1.5cm},clip]{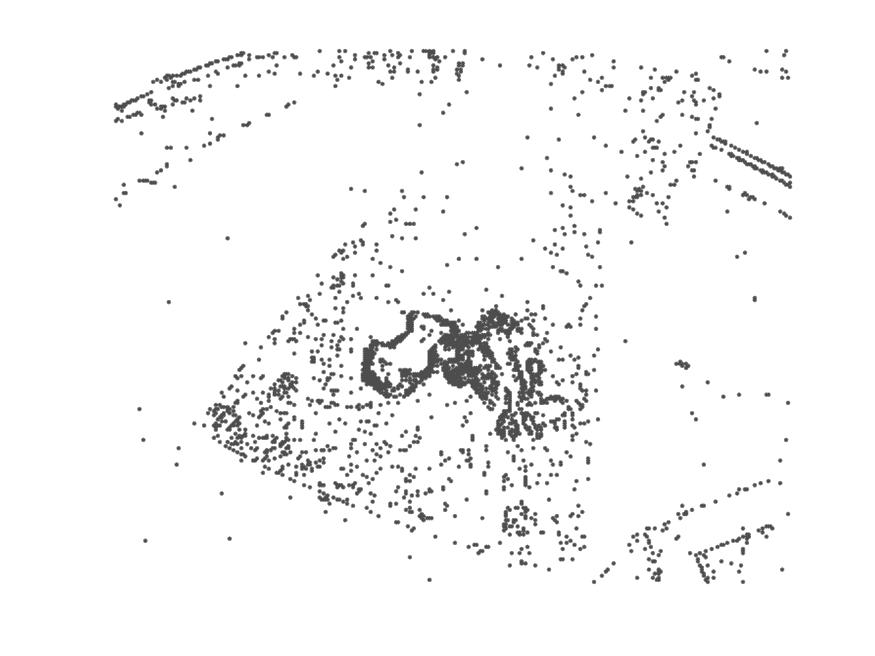}} 
&

\fbox{\includegraphics[width=\columnwidth,angle =0,trim={3.8cm 2.4cm 2.8cm 1.5cm},clip]{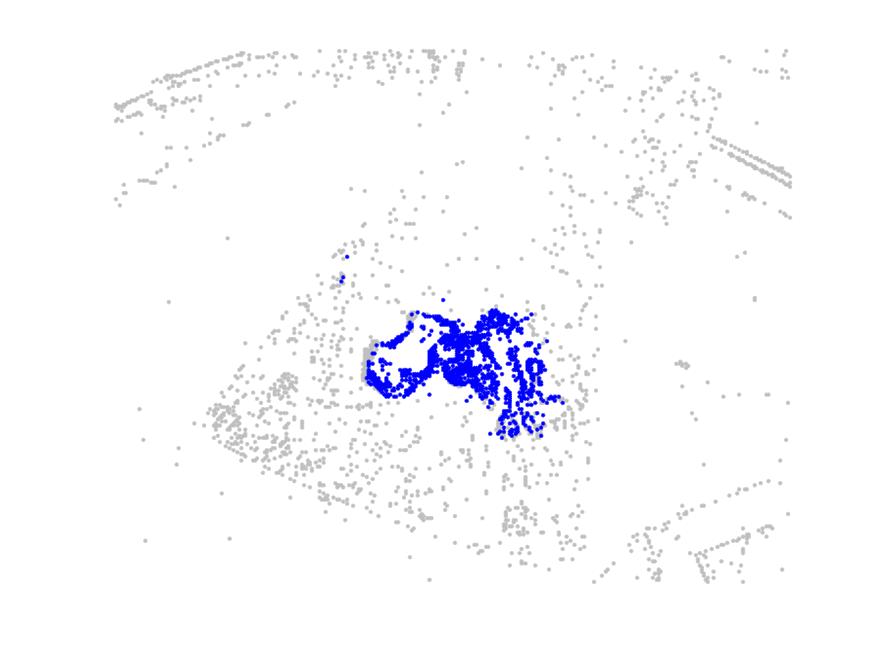}}&

\fbox{\includegraphics[width=\columnwidth, height=7cm,angle =0,clip]{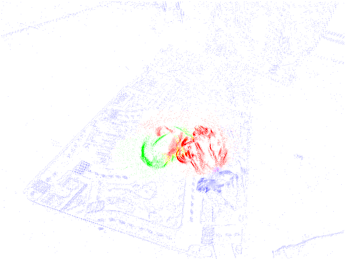}}


\\\\
\multicolumn{4}{c}{\centering\fontsize{30}{30}\selectfont { (B) EV-IMO - Table seq01 - two dynamic objects within the scene}}\\\\

\end{tabular}}
\end{adjustbox}
  \setlength{\belowcaptionskip}{-6pt}   

\caption{\textcolor{black}{Qualitative comparison with classical approach for event-based motion segmentation on EV-IMO testing sequences (Boxes and Table). Each sample presents (left to right) APS frames (for visualization only), raw event stream, Zhou et al. \cite{M10_Zhou21tnnls} segmentation labels, and our GTNN prediction, respectively.}}\label{qual_3}

\end{figure}

 Although Zhou et al. \cite{M10_Zhou21tnnls} is an offline approach that undergoes an initialization stage based on motion compensation, the proposed GTNN segmentation results are highly accurate and comparable with Zhou et al. \cite{M10_Zhou21tnnls} as depicted in Fig. \ref{qual_3}. In Boxes scene, one dynamic object, a toy car, traverses a textured carpet from left to right and we continuously segment the moving objects along the path, as shown in Fig. \ref{qual_3}-(A). On the other hand, in the Table scene, two dynamic objects, a toy plane and a car, move towards each other and then collide in the middle above the carpet, as shown in Fig. \ref{qual_3}-(B). When the dynamic objects collided, our model segmented them together as one object. Segmentation results have proved the robustness of the proposed algorithm in presence of background variations and multiple dynamic objects.

\begin{figure}[!bt]
\centering
\setlength{\fboxrule}{0.8pt}%
 \setlength{\belowcaptionskip}{-12pt}   

\centering
\begin{adjustbox}{width=0.47\textwidth}
  \centering

\begin{tabular}{cc|cc}

\centering\fontsize{30}{30}\selectfont{{Ground Truth}} & \centering\fontsize{30}{30}\selectfont{\textbf{GTNN (Ours)}}
&\centering\fontsize{30}{30}\selectfont{{Ground Truth}}&\fontsize{30}{30}\selectfont{\textbf{GTNN (Ours)}}
\\ \\

\fbox{\includegraphics[width=\columnwidth,trim={3.8cm 2.4cm 2.8cm 1.5cm},clip]{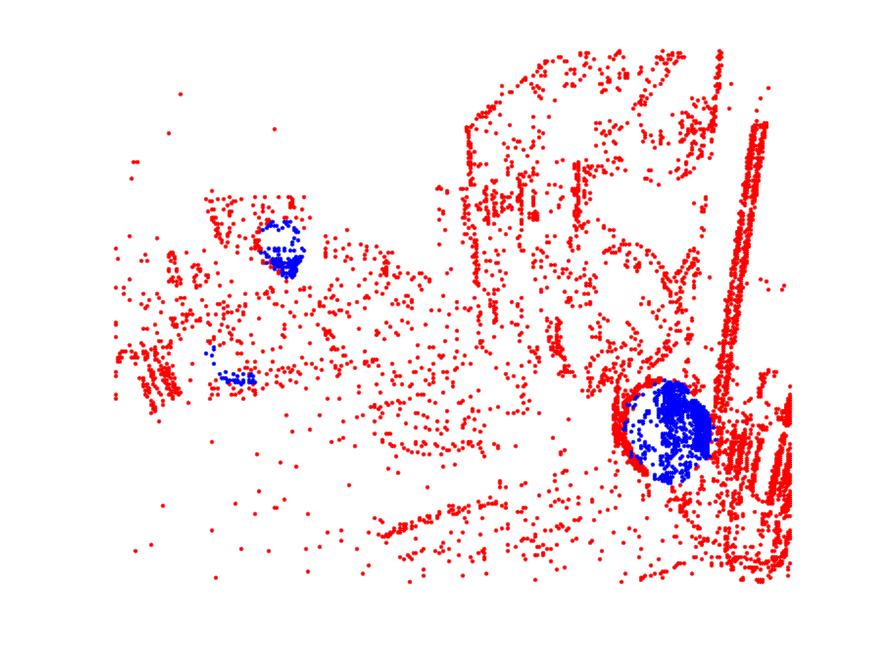}}

&
\fbox{\includegraphics[width=\columnwidth,trim={3.8cm 2.4cm 2.8cm 1.5cm},clip]{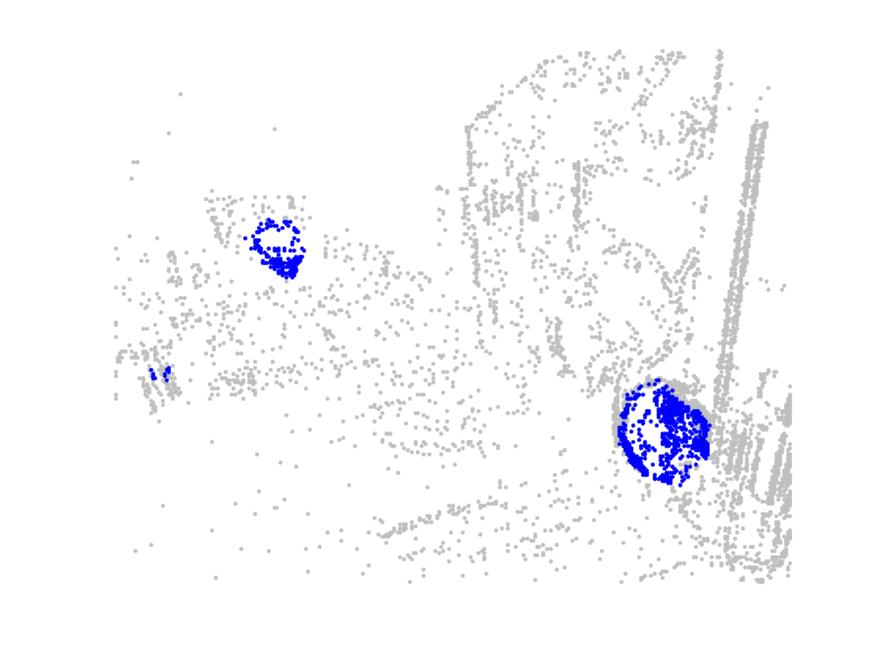}}
&
\fbox{\includegraphics[width=\columnwidth,trim={3.8cm 2.4cm 2.8cm  1.5cm},clip]{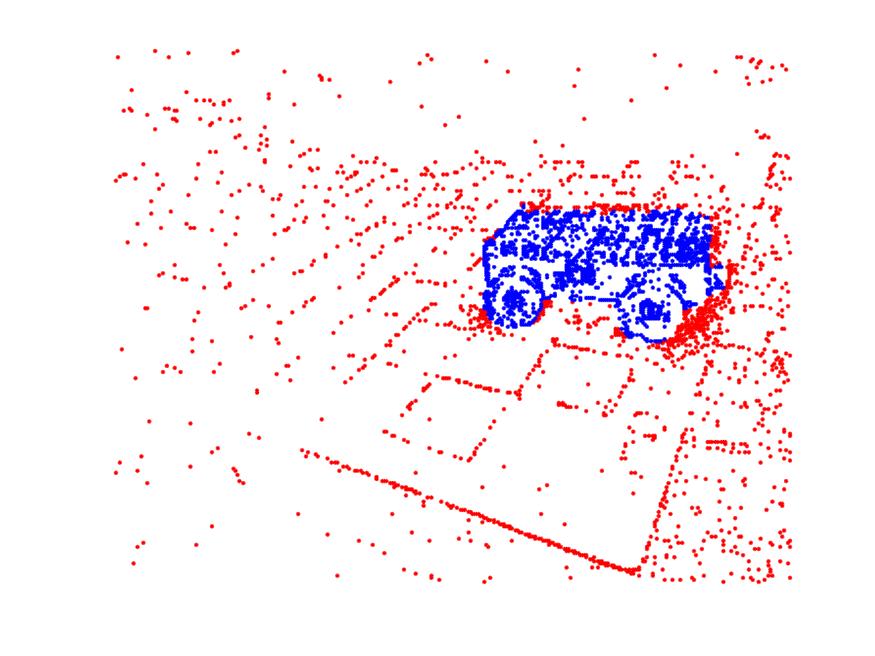}}

&
\fbox{\includegraphics[width=\columnwidth,trim={3.8cm 2.4cm 2.8cm 1.5cm},clip]{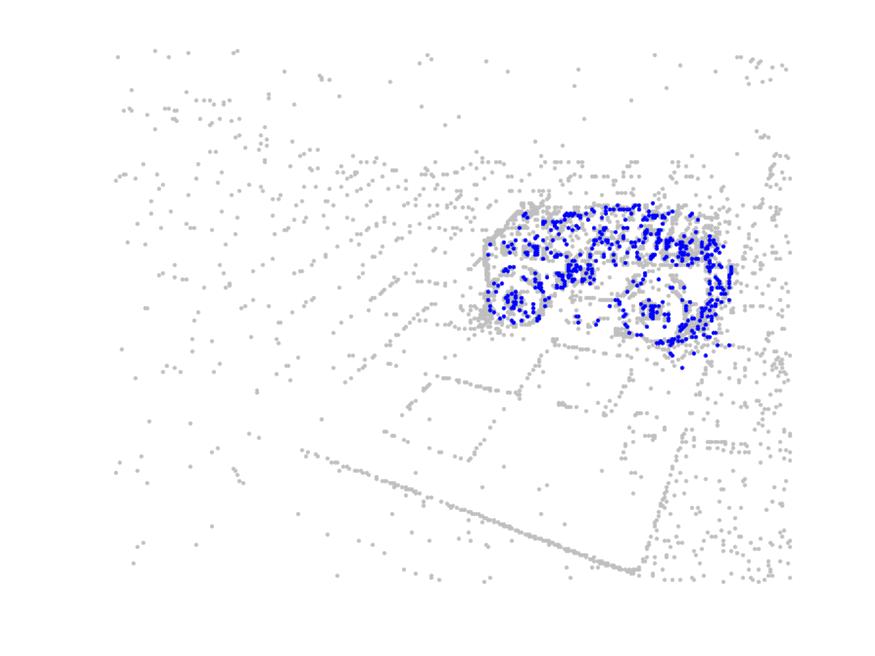}}
\\ \\
\multicolumn{2}{c}{\centering\fontsize{30}{30}\selectfont { (A) {MOD - Evaluation  - test\textunderscore seq }}} & \multicolumn{2}{c}{\centering\fontsize{30}{30}\selectfont { (B) {EV-IMO2 - Evaluation Set- seq\textunderscore 14\textunderscore 03 }}}\\\\

\scalebox{1}[1]{\fbox{\includegraphics[width=\columnwidth,height=7cm,angle =0,trim={3.8cm 2.4cm 2.8cm  1.5cm},clip]{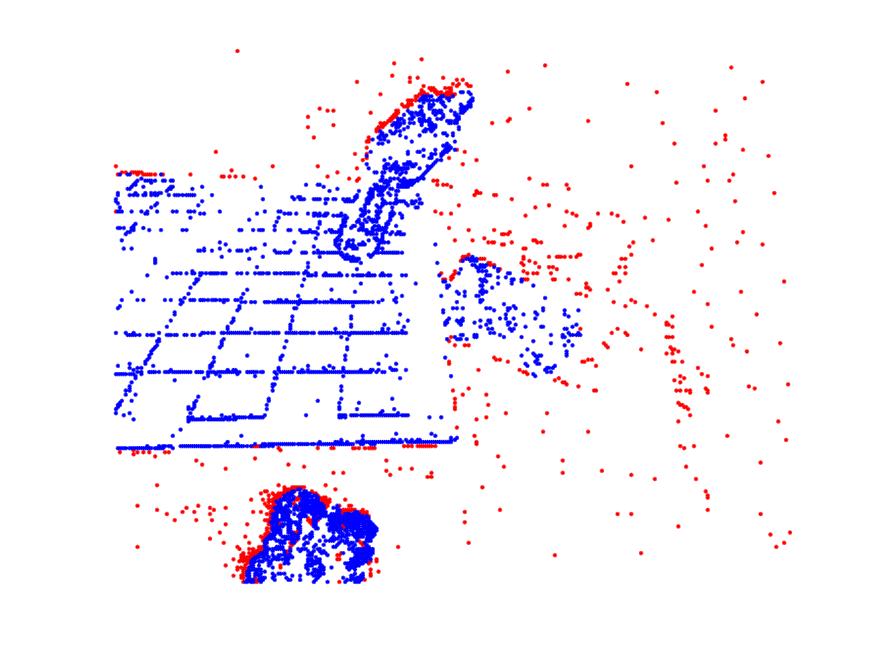}}}
&
\scalebox{1}[1]{{\fbox{\includegraphics[width=\columnwidth,height=7cm,angle =0,trim={3.8cm 2.4cm 2.8cm 1.5cm},clip]{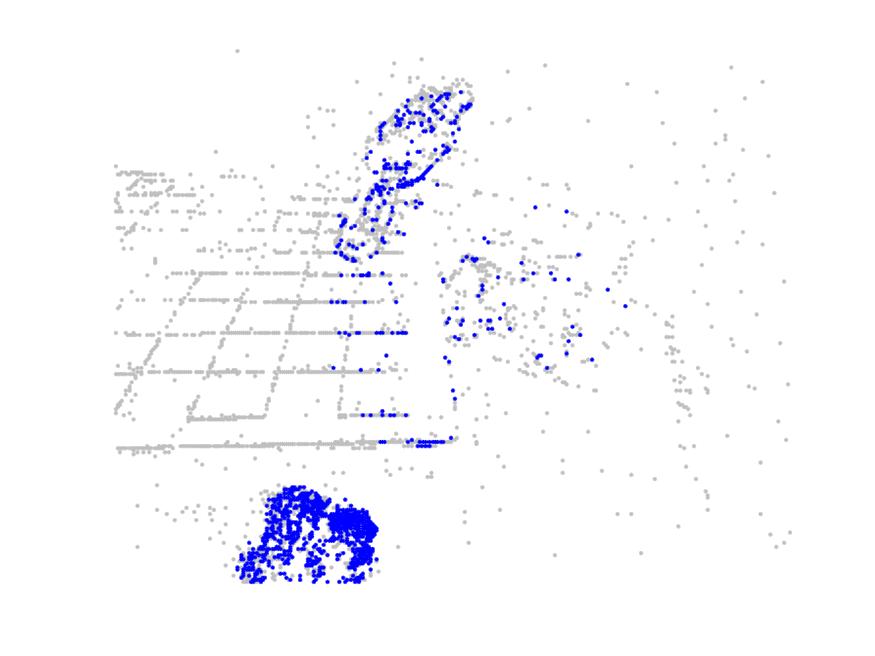}}}}

&

\fbox{\includegraphics[width=\columnwidth, height=7cm,trim={3.8cm 2.4cm 2.8cm 1.5cm},clip, angle=0]{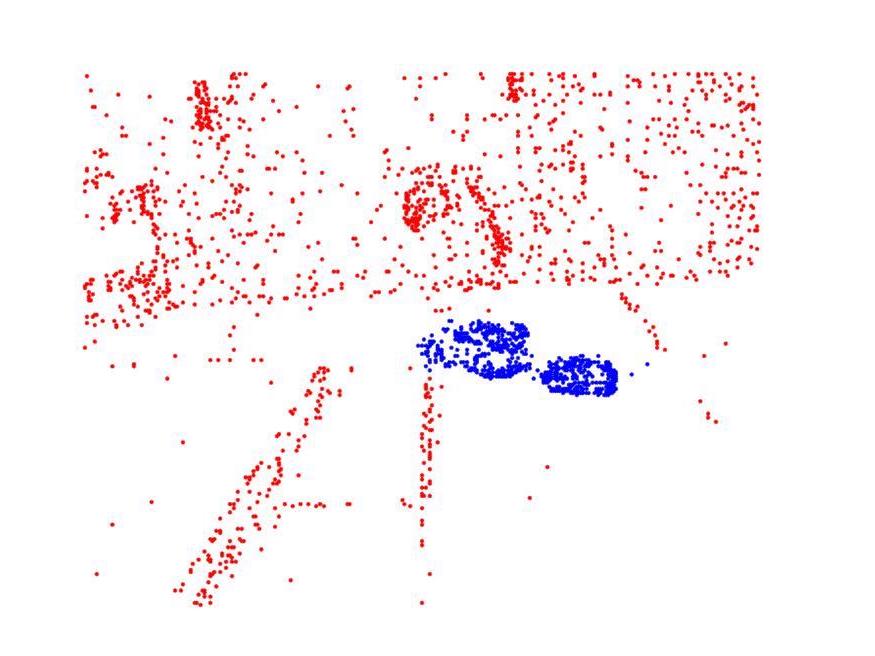}  }
&
\fbox{\includegraphics[width=\columnwidth,height=7cm,trim={3.8cm 2.4cm 2.8cm 1.5cm},clip, angle=0]{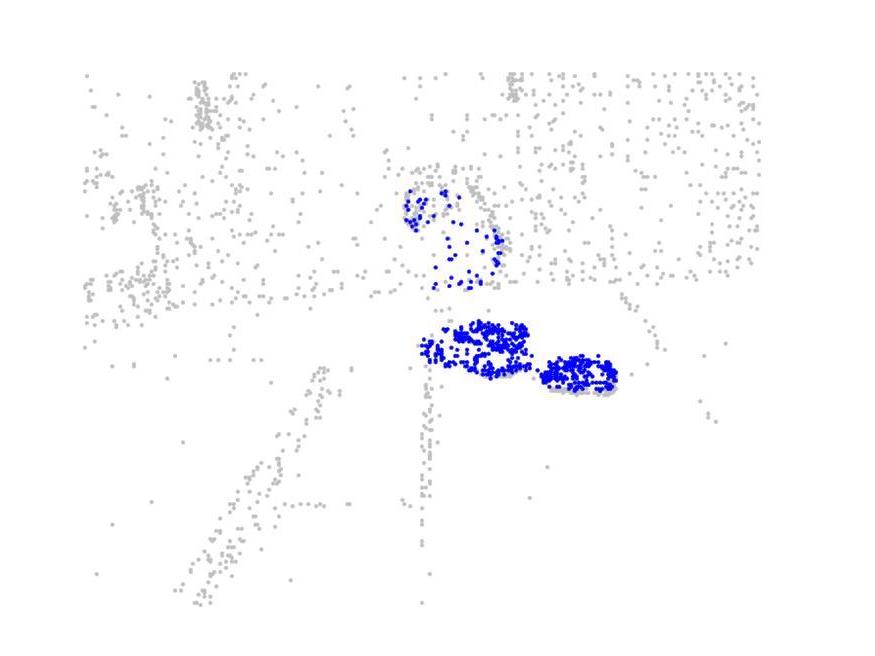}  }

\\ \\
\multicolumn{2}{c}{\centering\fontsize{30}{30}\selectfont{ (C) {EV-IMO2 - Evaluation Set - seq\textunderscore 15\textunderscore 02  }}}

&

 \multicolumn{2}{c}{\centering\fontsize{30}{30}\selectfont { (D) {Ours (EMS-DOMEL) dataset - Seq1}}} \\

\\

\\
\end{tabular}
\end{adjustbox}
 \setlength{\belowcaptionskip}{-9pt}   

\caption{\textcolor{black}{Qualitative evaluation of our GTNN algorithm on unseen testing simulation (MOD) and experimental (EV-IMO2 and our EMS-DOMEL) datasets. Our GTNN motion segmentation algorithm is able to identify dynamic objects within a variety of scenes and from various event camera resolutions (346$\times$260 and 640$\times$480), when the camera is in motion. }}\label{qual_2}
\end{figure}

\subsubsection{MOD, EV-IMO2 and our datasets (EMS-DOMEL)}

Fig. \ref{qual_2} provides additional qualitative results of our approach on the simulated MOD dataset and experimental EV-IMO2 dataset compared against approximate event labels estimated from the mask of dynamic objects, provided alongside the datasets. Note that these sequences are not fed to our GTNN model during training. As shown in Fig. \ref{qual_2}-(A), the approximate ground truth has some false labeling of the true dynamic objects (i.e. edges), however, our GTNN model predicts correctly the labels of the detected dynamic objects including their edges.  
Although, our model is trained with these approximate labels, i.e. including \textcolor{black}{some} false labels, the trained GTNN is able to exploit the spatiotemporal correlations of the events that belong to the moving object to correctly segment them from the background events. Fig. \ref{qual_2}-(B-C) presents the predictions of GTNN model when tested on scenes from the EV-IMO2 dataset. Our segmentation shows good performance compared to the ground truth. 

Further testing and analysis are done on our EMS-DOMEL recorded experimental dataset without any further training or fine-tuning of GTNN. GTNN demonstrated good segmentation and generalization capabilities. 
Lastly, we provide our segmentation results of a scenario where two toy cars with different sizes move from right to left and then collide with each other ({EMS-DOMEL-Seq1}). GTNN predictions continuously detect the moving objects (cars) while the camera is in motion, as shown in Fig. \ref{qual_2}-(D). 
Additional segmentation results can be found at  $<$\url{https://youtu.be/9z3Ik8V45Ms}$>$ and are also provided in the supplementary material.
To that end, the presented qualitative analyses have proven the capability of the proposed GTNN model to (1) cope with different camera parameters and various camera resolutions, and (2) generalize well to various dynamic scenes with multiple dynamic objects (moving at different speeds and in different directions) while the camera is in motion.

\subsection{GTNN Transferability Across Different Domains}

\textcolor{black}{GTNN is engineered in a way to facilitate its adaptability and generalizability, enabling its application across various domains and modalities. To tailor it for specific tasks, further refinement is often necessary. For instance, Sanket et al. \cite{sanket} employed GTNN in the domain of event-based panoptic segmentation, by means of a transfer learning approach. They fed a 3D event stream graph into the GTNN, facilitating the exchange and nonlinear transformation of features to extract both local and global spatiotemporal relationships among graph nodes. The processed information is then directed to a final layer, which is specifically adjusted for the panoptic segmentation task. Hence, training was conducted for the final layer only, leaving the parameters of the early layers in the GTNN architecture unchanged, as demonstrated in \cite{sanket}.}

\textcolor{black}{Beyond vision-related tasks, GTNN's adaptability extends to non-visual tasks as well. This is intrinsic to its design framework, and is achieved by modifying the nodes and edges within the 3D graph structure to encapsulate the unique aspects of the problem being addressed. For instance, in non-vision-related tasks, the graph elements could be redefined to represent different data types or relationships, showcasing the GTNN's broad generalization capabilities. This adaptability underscores the potential of GTNN to serve as a base architecture for a wide range of applications, provided that appropriate adjustments are made to align with the specific requirements of each task.}

\subsection{Limitations}
\subsubsection{Segmentation Challenges in Dynamic Environments}
\textcolor{black}{Rigorous evaluations are conducted on the publicly-available datasets as well as our recorded experiments demonstrating the superiority of the GTNN to segment dynamic objects (Sections \ref{sec:qaun} and \ref{seq:qualitative}). Our evaluations indicated that GTNN struggles to differentiate dynamic objects from the background in scenarios where the camera and object speeds are similar, often misclassifying all events as background. This limitation becomes more pronounced in environments with minimal relative motion between the camera and the object, leading to inaccurate foreground-background segmentation. Furthermore, GTNN encounters difficulties during rapid camera rotations, especially in situations where the scene is cluttered with numerous background events. In these situations, the model's effectiveness in recognizing and segregating smaller or distant moving objects is reduced. Rapid camera movements exacerbate the problem by flooding the algorithm with an excessive event stream. This issue is evidenced by our observed reduction in performance for the 'EV-IMO Boxes' scenarios, as detailed in Table \ref{tab:Iou_spike}. Such scenarios pose a substantial challenge for GTNN in consistently recognizing and tracking objects. It is worth mentioning that our GTNN yet demonstrates outstanding performance in some 'EV-IMO Boxes' evaluation sequences, as presented in Fig. \ref{qual_3}.}

\subsubsection{Computational Time Analysis}  
\begin{figure}[!t]
\centering

 \includegraphics[width=0.440\textwidth]{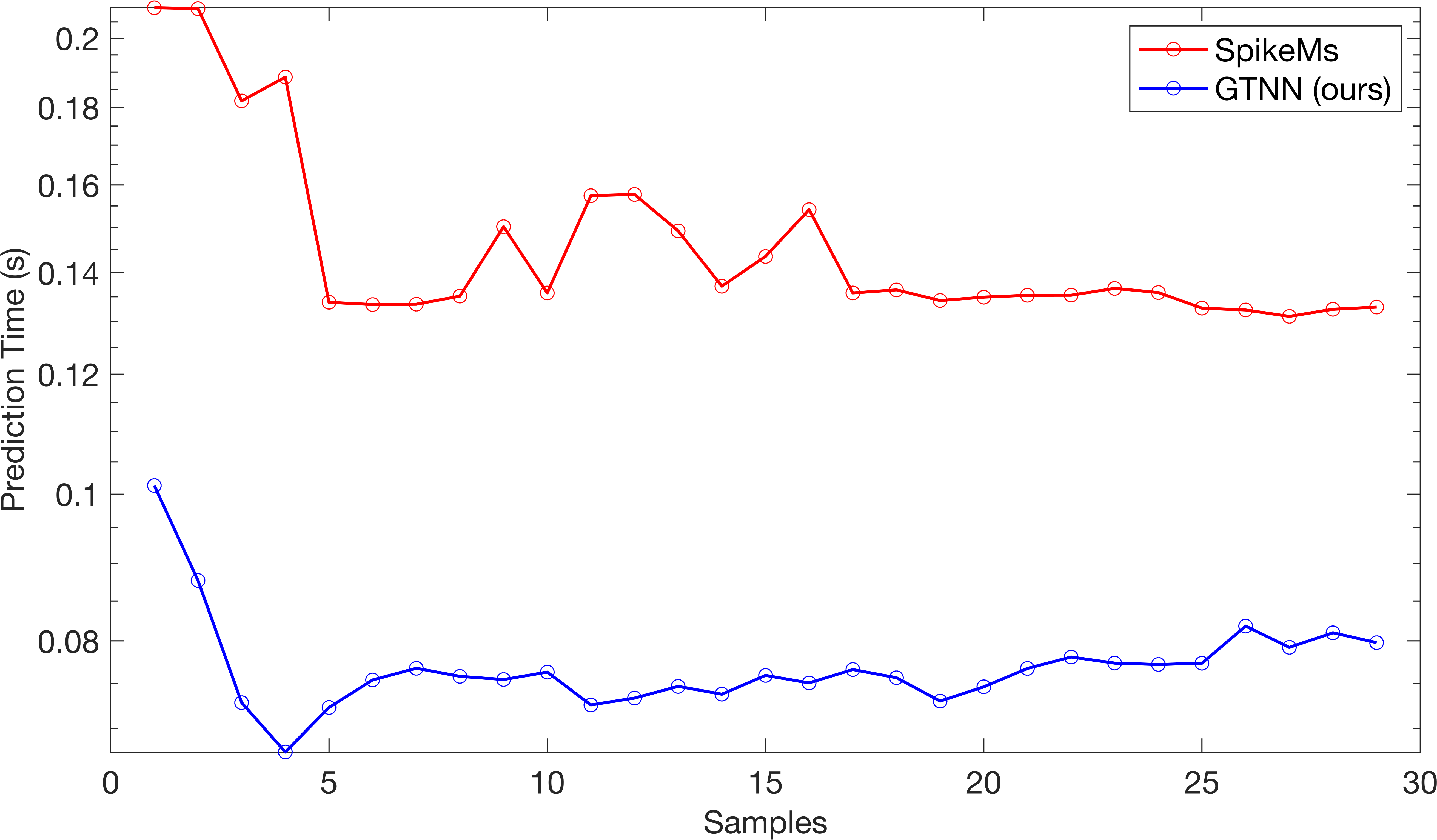}
  \setlength{\belowcaptionskip}{-8pt}   

 \caption{Forward time in seconds to perform motion segmentation for 10ms time window of event stream using the proposed approach and SpikeMS \cite{SpikeMS}}
  \label{fig:computation_time}
\end{figure}

\textcolor{black}{In this section, analysis of computational time required to process input event streams by the proposed GTNN algorithm is presented and compared to SpikeMS \cite{SpikeMS}. 
Note that timing analysis was carried out on a Dell Desktop Computer with Intel(R) Xeon(R) W-2145@2.70GHz×8 and two Nvidia Quadro RTX 6000 GPUs. 
A set of 30 event graphs, each spanning a 10ms time window was selected from the Boxes sequence of {EVI-MO} dataset to conduct the time analysis. The computational time analysis of the proposed algorithm compared to SpikeMS \cite{SpikeMS} was done in terms of the forward prediction time as shown in \textcolor{black}{Fig. \ref{fig:computation_time}}. It is observed that the time needed to process each event graph was shorter using our proposed approach than SpikeMS \cite{SpikeMS} achieving a double speedup of processing time. }

\textcolor{black}{{There} is yet more room for improvement to expedite the runtime performance of the proposed approach to fulfill the real-time performance requirement and achieve the primary goal of the event-based motion segmentation. Therefore, to achieve real-time performance, processing a group of events within a selected time window should be performed before receiving the next event graph, i.e. in our case processing should be done in less than 10ms. This could be achieved by optimizing the complexities of the GTNN architecture via knowledge distillation methods \cite{wang2021knowledge}. More particularly, a reduced network architecture (called a student model) is trained in parallel with the complex network (called a teacher model) which has enough nonlinear capacity to handle the problem at hand. The student model, when trained alone, does not have the capability to map the nonlinearities in the input and output dataset to provide the required performance, however, with the student-teacher parallel training methods, this can be achieved. Alternatively, to reduce the runtime, a simpler network than the proposed GTNN architecture is suggested to be trained using knowledge distillation framework. These areas for improvement will be addressed in the future. }

\raggedbottom

\section{Conclusions}\label{sec:conclusion} 
In this work, we presented the first learning-based graph transformer neural network (GTNN) algorithm tackling a large-scale problem in computer vision using dynamic vision sensors. Our GTNN algorithm is developed to infer spatiotemporal patterns of the acquired event streams and perform event-based motion segmentation accordingly. More specifically, the algorithm reveals the motion dynamics of the camera and the scene and decides whether the incoming events represent foreground (due to moving objects) or background (due to camera motion) event data.

The proposed GTNN successfully operates on event streams without any initialization stage nor prior knowledge in terms of scene geometry, motion patterns, or number of independent dynamic objects. This is attributed to the fact that the adopted graph structure of the input data exploits the spatiotemporal patterns between the events, then infers them in the global context of the scene. Event streams are processed in their asynchronous form which preserves their temporal attributes, i.e. without projecting them into 2D frames. Such operation is carried out in the point transformer with self-attention mechanism, transition up and transition down layers where inputs are fed as 3D event graphs which could be of variable sizes.

GTNN algorithm was trained using the proposed effective scheme on three publicly available synthetic and real-work sequences, MOD, EV-IMO, and EV-IMO2. The effective training scheme uses a portion of the extensive training datasets at every iteration (epoch) for tuning the network weights. The proposed effective scheme has reduced the training time and expedited convergence. The proposed GTNN algorithm has outperformed the SOTA learning-based motion segmentation \cite{SpikeMS} and classical methods \cite{M10_Zhou21tnnls,M9_s128_Mitrokhin_2018} in all the testing scenarios with at least 4.5\% higher detection rate and 9.5\% accuracy (\textit{IoU}\%) on testing sets. Moreover, the forward prediction time is 50\% less compared to SOTA learning-based approach. Furthermore, qualitative results have proven the superiority of the proposed algorithm to both the learning-based motion segmentation approach \cite{SpikeMS} and the classical approach proposed in \cite{M10_Zhou21tnnls}.

Our GTNN model was also tested on our experimental dataset which was not exposed to the network during training. A novel offline event labeling technique, referred to as DOMEL, is proposed to label the recorded dataset, which involves event streams capturing moving objects when the camera is in motion. GTNN is able to successfully segment the moving objects from the background, despite the fact that the data is recorded under conditions different than those of the training data; a variety of scenes, multiple objects moving with various speeds and in random paths, and unknown camera motions with various sensor resolutions.
 This work in addition to our previous research \cite{AlkendiY} have unlocked the potential of using graph transformers neural networks on vision-based navigation modules with event camera. In the future, we plan to demonstrate the significance of our proposed event-based motion segmentation algorithm by integrating it with a dynamic object avoidance module to perform safe navigation in unknown dynamic environments.

\section*{Acknowledgements}
This research publication was funded by the Khalifa University of Science and Technology under Award No. RC1-2018-KUCARS, and by the Advanced Research and Innovation Center (ARIC), which is jointly funded by Khalifa University of Science and Technology and STRATA Manufacturing PJSC (a Mubadala company), grant number 8436010.
The author(s) wish to acknowledge the contribution of Khalifa University's high-performance computing and research computing facilities to the results of this research. The author(s) also thank Mohammed Salah and Oussama Abdul Hay for their help during data collection.

\bibliographystyle{IEEEtran}
\bibliography{bibtex/bib/refMS}

 \appendices

\end{document}